\documentclass[10pt,twocolumn,letterpaper]{article}
\usepackage[pagenumbers]{cvpr} 

\usepackage{cuted}      
\usepackage{currfile}
\usepackage{caption}
\usepackage{amsmath,amssymb}
\usepackage{booktabs}
\usepackage{multirow}
\usepackage{graphicx}
\usepackage{xspace}

\usepackage{wrapfig}

\newcommand{\wish}{WISH\xspace}

\DeclareMathOperator*{\argmax}{arg\,max}

\usepackage{afterpage}

\definecolor{cvprblue}{rgb}{0.21,0.49,0.74}
\usepackage[pagebackref,breaklinks,colorlinks,allcolors=cvprblue]{hyperref}

\def\confName{CVPR}
\def\confYear{2027}

\title{WISH: Word-Informed Selection of Heads for Vision-Language Explainability}

\author{Karim Wahba$^{1}$ \qquad Madhav Iyengar$^{1}$ \qquad Walid Bousselham$^{2}$ \qquad Hilde Kuehne$^{1}$\\
$^{1}$University of T\"ubingen \qquad $^{2}$Google DeepMind
}

\begin{document}
\maketitle
\begin{abstract}
Vision-language models (VLMs) are being increasingly employed for vision-language tasks. 
However, understanding the visual evidence behind each generated word remains a complex task. 
A key challenge in VLM explainability is choosing where in the model to extract this evidence. 
Existing methods address this by using final-layer visual-token activations or aggregating attention gradients across layers and heads. 
However, focusing on the final layer can miss intermediate grounding signals, while  gradient weighting can misrepresent how much each head contributes. 
In contrast, we introduce \wish (Word-Informed Selection of Heads), a word-level explainability method that adaptively selects heads across layers to construct word-level visual explanations. \wish ranks the heads based on an estimate of how much blocking each head's visual attention would lower the word's layer logit. It subsequently selects the highest scoring heads and constructs heatmaps by spatially filtering and aggregating their gradient maps. Across six VLMs, attention knockout experiments show that \wish-selected heads have a greater effect on word prediction than those selected using five alternative criteria. Spatial map evaluations on ImageNet, VQAv2, PASCAL VOC, and COCO Captions further demonstrate improved object localization and stronger perturbation-based performance in most comparisons with existing methods. Finally,  we find evidence that when VLMs generate function words, e.g., articles, their explanation heatmaps already highlight the associated objects in the image. We term this behavior ``look-ahead grounding'' and propose two evaluation metrics to quantify it. Project page: \url{https://karim-wahba.github.io/wish}.
\end{abstract}

\begin{figure}[!t]
\centering
\includegraphics[width=\linewidth]{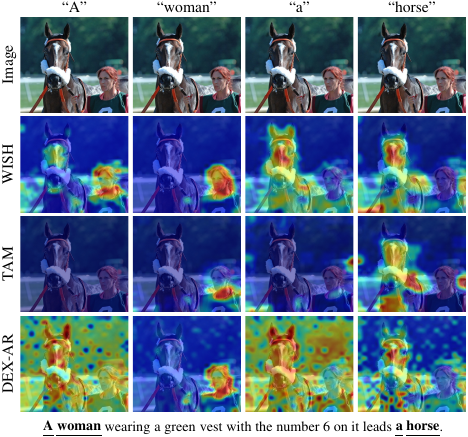}
\vspace{-6mm}
\caption{Function-word and noun explanations: \wish's maps for “A” and “a” highlight regions of the woman and horse, respectively, before they are generated, while TAM's are nearly empty and DEX-AR's are spatially diffuse.}
\label{fig:teaser}
\vspace{-6mm}
\end{figure}

\section{Introduction}
\label{sec:intro}

Vision-language models have demonstrated remarkable capabilities in interpreting complex visual scenes, reasoning about spatial relationships, and understanding videos~\cite{qwen3-vl-2025,wang-internvl35-2025,clark-molmo2-2026,google-gemini31pro-2026}.  
Despite these advances, VLMs tend to generate visually unsupported descriptions, sometimes relying on language priors rather than the presented visual information~\cite{bai-mllm-hallucination-survey-2024,lietzow-prior-override-2026}. 
Explainability methods therefore try to understand the visual basis of VLM predictions, helping users assess outputs and developers diagnose failures~\cite{adebayo-debugging-tests-2020,bansal-whole-exceed-parts-2021}.

A central challenge in understanding this visual grounding of an output is identifying which parts of the model the evidence arises from.
This matters because different model layers and attention heads contribute unequally to each word prediction.
Schaumlöffel et al.~\cite{schaumloeffel-object-localization-2026} show that a small subset of attention heads contributes to object classification and localization. 
Similarly, related studies have identified attention heads specialized for visual grounding and retrieval~\cite{kang-localization-heads-2025,li-visual-retrieval-heads-2026}.
Recent word-level VLM explainability methods address the problem of what parts of the model to use by constructing 2D heatmaps extracted from the last layers or aggregated across multiple layers and attention heads. 
TAM~\cite{li-token-activation-map-2025} constructs heatmaps from final-layer visual-token activations, while DEX-AR~\cite{bousselham-dexar-2026} aggregates attention gradients across layers and heads, weighting them according to their relative visual and textual relevance.
However, relying only on the final layer can miss grounding signals formed in intermediate layers, while heuristic gradient weighting across all heads does not reflect each head's contribution to predicting the target word.

In contrast to these works, this paper introduces \wish (Word-Informed Selection of Heads), an explainability method that constructs word-specific visual explanations by first identifying relevant attention heads according to their estimated predictive contribution. 
Namely, \wish considers the importance of each head based on the effect that knocking out its visual attention would have on the target word's prediction, as measured by attention knockout.
However, since evaluating each score via attention knockout at inference time is computationally impractical, we approximate the impact of each head's visual attention score via the first-order Taylor expansion of the attention knockout formulation. 
We use the resulting score to rank the respective heads and use an adaptive selection method to select the finally used heads. 
To construct the heatmap, we combine elements from DEX-AR and TAM, using attention-gradients as proposed by DEX-AR to avoid attention sinks~\cite{bousselham-dexar-2026}, and denoising filters as proposed by TAM to filter spurious outliers~\cite{li-token-activation-map-2025}. 
Distinct from the two methods, \wish applies this process only to
the preselected heads of each word and aggregates their maps with a rank-weighted median filter.

Our evaluation spans six VLMs, 
LLaVA-1.5-7B~\cite{liu-llava15-2024}, BakLLaVA~\cite{skunkworks-bakllava-2023}, Qwen3-VL-8B~\cite{qwen3-vl-2025}, GLM-4.6V-Flash~\cite{vteam-glm45v-2025}, LLaVA-OneVision-1.5-8B~\cite{an-llava-onevision-15-2025}, and LLaVA-OneVision-2-8B~\cite{an-llava-onevision-2-2026},
and four benchmarks covering
perturbation and segmentation on ImageNet, VQAv2, PASCAL VOC, and COCO Captions. 
We further evaluate head selection across different selection criteria using attention knockout interventions. 
These experiments show that \wish identifies influential heads and produces visual explanations that are competitive with or improve upon existing methods.

Finally, we address the problem that current VLM explainability methods  mainly focus on explaining nouns that represent objects in the image. 
Studies in NLP have shown that LLMs encode information about upcoming tokens before generating them~\cite{pal-future-lens-2023,men-lookahead-planning-2024}. 
Using \wish, we find evidence that a related phenomenon exists in VLMs as well. Explanations of function words, such as the article in the phrase ``a horse'', already highlight the image regions of the objects whose nouns are yet to be generated (see
Fig.~\ref{fig:teaser}).
We term this behavior \emph{look-ahead grounding} and quantify it by introducing two complementary metrics. First, \textit{Function-Word Perturbation} (FWP)  measures whether perturbing a function word's image region affects its associated noun prediction more than perturbing the regions of other function words. Second, \textit{Function-Word Segmentation} (FWS) compares the alignment of a function word's explanation with associated objects against other objects named in the caption. Our findings extend the study of look-ahead behavior in language models to the visual grounding of upcoming words in VLMs.

Our contributions are summarized as follows:
(1) We introduce \wish, a VLM explainability method that leverages a head-selection mechanism based on a first-order approximation of the attention-knockout effect of the heads' visual tokens. 
(2) We propose to construct heatmaps only from the selected heads through a combination of attention gradients and denoising filters.
(3) Using \wish, we show that VLMs exhibit \emph{look-ahead grounding} in function words. We introduce two metrics, testing whether a function word's map affects its noun more than other function words' maps and overlaps the object it introduces.

\section{Related work}
\label{sec:related}
\noindent\textbf{Word-Level Explainability Methods.}
Visual explanation methods aim to identify the visual evidence in image regions behind the model's outputs~\cite{simonyan-saliency-2014,zhou-cam-2016,selvaraju-gradcam-2017}. For convolutional neural networks~\cite{krizhevsky-alexnet-2012} and vision transformers~\cite{dosovitskiy-vit-2021}, these methods derive spatial attributions from activations~\cite{zhou-cam-2016}, gradients~\cite{selvaraju-gradcam-2017,sundararajan-integrated-gradients-2017}, relevance propagation~\cite{bach-lrp-2015,achtibat-attnlrp-2024}, or attention~\cite{abnar-rollout-2020,chefer-generic-attention-2021}. However, their application to VLMs is complicated by the interaction between visual inputs and preceding text during autoregressive generation. Recent word-level explainability methods developed for VLMs adapt to this setting by constructing heatmaps using either internal model signals or image perturbations. The former examine activations and gradients within the model, whereas the latter measure how predictions change when parts of the image are modified. Among methods using internal signals, TAM~\cite{li-token-activation-map-2025} constructs token-specific activation maps by projecting final-layer visual-token representations onto vocabulary weights, subtracting estimated interference from preceding tokens, and applying spatial filtering. DEX-AR~\cite{bousselham-dexar-2026} aggregates attention-gradient maps across layers, dynamically weighting heads according to the relative strength of their visual and textual gradients. In contrast, EAGLE~\cite{chen-eagle-2026} adopts a black-box approach, identifying image regions sufficient or necessary for generating selected tokens through greedy perturbation search. While being highly model agnostic, its search requires repeated model evaluations, with quadratic complexity in the number of image regions. 
\wish uses internal model signals, but unlike TAM and DEX-AR, it explicitly selects heads by their estimated effect on the predicted word and constructs the explanation from these heads only.

\noindent\textbf{Attention-Head Importance.} Transformers process information through multiple attention heads, but heads differ in their importance and role. Early studies of transformer-based language models showed that many heads could be removed without substantially degrading performance~\cite{michel-sixteen-heads-2019,voita-heads-2019}. Studies of VLMs specifically have identified heads specialized in locating objects and retrieving visual information~\cite{kang-localization-heads-2025,li-visual-retrieval-heads-2026}. These studies identify heads' relevance for certain tasks or models by examining their attention patterns. However, whether attention weights faithfully reflect which inputs drive a model’s prediction remains debated~\cite{jain-attention-not-explanation-2019,wiegreffe-attention-not-not-2019}.  Studies show that attention can identify visual regions necessary for prediction in some settings, but can also fail to reflect the model's reliance on attended information or concentrate on visually irrelevant tokens known as attention sinks ~\cite{song-etal-2026-visual,kang-visual-attention-sink-2025}. A complementary way to assess this dependence is through attention knockout, which intervenes by blocking selected attention connections and measures their effect on model predictions~\cite{geva-dissecting-recall-2023,kaduri-whats-in-the-image-2025}. 
However, since exhaustive knockout is expensive to run, gradient-based approximations are common. Michel et al.~\cite{michel-sixteen-heads-2019} rank heads by the gradient of the loss with respect to a head mask, and Attribution Patching ~\cite{kramar2024atp} recomputes the attention softmax before linearizing to avoid false negatives under saturated attention.
Building on this approach, WISH provides a word-specific score for ranking heads by estimating the effect of removing a head's positive visual attention on the target word's layer logit.

\noindent\textbf{Look-Ahead Behavior in Language Generation.}
Although language models generate tokens autoregressively, their internal representations can contain information about words that have yet to be generated. Pal et al.~\cite{pal-future-lens-2023} show that future tokens can be predicted from intermediate representations, while subsequent work investigated look-ahead mechanisms in language-model planning~\cite{men-lookahead-planning-2024}. These studies examine linguistic representations, leaving open whether VLMs ground information about upcoming words in specific image regions. This question is particularly relevant to function words, which are assumed to lack direct visual evidence in the input image, even when they precede words that refer to visible objects. For instance, TAM~\cite{li-token-activation-map-2025} introduces a metric that evaluates function word heatmaps, penalizing maps that highlight image regions. WISH challenges this assumption by examining whether function-word explanations exhibit look-ahead visual grounding.
\section{Method}
\label{sec:method}

\begin{figure*}[t]
    \centering
    \includegraphics[width=\textwidth]
    {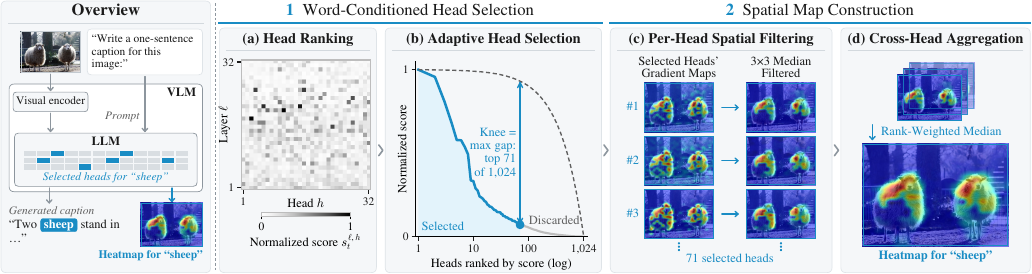}
    \vspace{-5mm}
    \caption{\wish explains the word ``sheep'' in the caption that LLaVA-1.5~\cite{liu-llava15-2024} generates for the image. \textbf{(a)}~Every attention head is scored by the head-ranking score $s_t^{\ell,h}$ (Eq.~\ref{eq:wish-score}). \textbf{(b)}~The knee of the sorted scores keeps the 71 highest-ranked of the model's 1,024 heads as the selected set $S_t$. \textbf{(c)}~The gradient map of each selected head is filtered with a $3\times3$ median; the maps of the three highest-ranked heads are shown. \textbf{(d)}~A rank-weighted median combines the filtered maps into the heatmap $M_t$ for ``sheep''.}
    \label{fig:method-overview}
    \vspace{-4mm}
\end{figure*}

\wish produces a heatmap for each generated word of an autoregressive VLM in two stages (Fig.~\ref{fig:method-overview}). First, it ranks attention heads and adaptively selects a subset. Second, it spatially filters the selected heads' attention-gradient maps and aggregates them across the heads into a final heatmap.

\subsection{Word-Conditioned Head Selection}
\label{sec:selection}

To rank attention heads, \wish estimates how blocking each head's visual attention would affect the target word's layer logit. We first introduce attention and the logit lens, which provide the quantities needed for this estimate.

\noindent\textbf{Preliminaries.} Autoregressive VLMs typically combine a visual encoder with a large language model (LLM). 
The LLM has $L$ transformer layers, each with $H$ attention heads, and generates tokens conditioned on the image, prompt, and generated tokens. For head $h$ in layer $\ell$, let $\mathbf{Q}^{\ell,h}$ and $\mathbf{K}^{\ell,h}$ be the query and key projections of the input token representations, respectively,
and $d_k$ the key dimension per head. The pre-softmax attention scores and normalized attention weights at each head are
\begin{equation}
\mathbf{e}^{\ell,h}
=\frac{\mathbf{Q}^{\ell,h}{\mathbf{K}^{\ell,h}}^{\top}}{\sqrt{d_k}},
\qquad
\mathbf{A}^{\ell,h}=\operatorname{softmax}\!\big(\mathbf{e}^{\ell,h}\big).
\label{eq:attention}
\end{equation}
We denote the attention weight in layer $\ell$, head $h$ from query position $i$ to key position $j$ by $A^{\ell,h}_{i,j}$.

\noindent\textbf{Logit Lens.} The logit lens~\cite{nostalgebraist-logit-lens-2020} formulation proposes a way to compute logits for intermediate layers in LLMs. 
Each layer $\ell$ maintains tokens $\mathbf{z}_i^\ell\in\mathbb{R}^d$ at sequence position $i$, where $d$ is the token dimension. During a generation step $t$, the token of the final-layer ($\ell=L$) at the last position ($i=-1$) is projected from the token $z_{-1}^{L}$ to the vocabulary space $\mathcal{V}$ via an unembedding matrix $\mathbf{W}_{\mathrm{U}}\in\mathbb{R}^{|\mathcal{V}|\times d}$ to predict the target word $y_t$. The logit lens~\cite{nostalgebraist-logit-lens-2020} applies the same projection to tokens in intermediate layers ($\ell \neq L$), yielding the layer logit formulation for $y_t$:
\begin{equation}
\hat{o}^{\ell}(y_t)=
\left[\mathbf{W}_{\mathrm{U}}\operatorname{Norm}(\mathbf{z}_{-1}^{\ell})\right]_{y_t},
\label{eq:logit-lens}
\end{equation}
where $\operatorname{Norm}$ denotes the final normalization. This provides an intermediate representation that can be used to measure the relevance of the layer $\ell$ to the prediction of $y_t$.

\noindent\textbf{Attention Knockout.}
Since attention heads vary in their relevance to word generation, we aim to rank them by the effect of their visual attention on the target word's prediction. 
To quantify this effect, we consider attention knockout interventions~\cite{geva-dissecting-recall-2023,kaduri-whats-in-the-image-2025}, a technique that sets the normalized attention scores of specific tokens to zero by setting  $\mathbf{e}^{\ell,h}=-\infty$. This blocks selected heads from attending to the token. The relevance of this specific token can then be defined by the decrease in the target word's layer logit as measured by the layer logit formulation (Eq.~\ref{eq:logit-lens}). 
Hence, for a head $h$ in layer $\ell$, we define the attention knockout score $\Delta^{\ell,h}$ for the visual token positions $R \subset\mathcal{I}$, as:
\begin{equation}
\Delta^{\ell,h}(R)
=
\hat{o}^{\ell}(y_t)
-
\hat{o}^{\ell,\setminus(h,R)}(y_t),
\label{eq:attention-knockout-score}
\end{equation}
where $\hat{o}^{\ell,\setminus(h,R)}(y_t)$ is the layer logit after blocking the head's attention to $R$. 

\noindent\textbf{Attention Knockout Approximation.} Computing this quantity for every head and every target word requires separate recomputations. We therefore approximate the knockout-induced logit change using a first-order Taylor expansion with respect to the attention weights. For notational simplicity, since the intervention is restricted to the predicting query position ($i=-1$) we write $A_k^{\ell,h}=A_{-1,k}^{\ell,h}$. 
Let $\mathcal K$ denote the set of key indices accessible from the predicting row $i=-1$, including the visual-token and the text-token indices. 
Let $G_k^{\ell,h}=\partial\hat{o}^{\ell}(y_t)/\partial A_k^{\ell,h}$ denote the attention-gradient for each accessible key $k\in\mathcal K$ and $\bar G^{\ell,h}=\sum_{k\in\mathcal K}A_k^{\ell,h}G_k^{\ell,h}$ the attention-weighted mean gradient. The derived first-order approximation is
\begin{equation}
\Delta^{\ell,h}(R)
\approx
\frac{\sum_{j\in R}A_j^{\ell,h}
\left(G_j^{\ell,h}-\bar G^{\ell,h}\right)}
{1- \sum_{j\in R}A_j^{\ell,h} }.
\label{eq:wish-knockout}
\end{equation}
The centered gradients account for attention redistribution, while the denominator captures softmax renormalization. The sum $\sum_{j\in R} A_j^{\ell,h}$ is the total attention weight on the visual tokens $R$ being blocked. It is less than $1$ because some attention weight is assigned to text tokens, which are outside $R$. The complete derivation is provided in Supp.~\ref{app:knockout}.

\noindent\textbf{Head Ranking via Token Scoring.} We use this knockout approximation to score and rank the heads on visual tokens whose removal is expected to decrease the layer logit and consequently increase $\Delta^{\ell,h}$. 
To identify these tokens, for each visual token $j\in\mathcal I$, we define a token's score as $C_j^{\ell,h}=A_j^{\ell,h}(G_j^{\ell,h}-\bar G^{\ell,h})$ and evaluate the approximation on tokens with positive scores $\lfloor C_j^{\ell,h} \rfloor_+$, where $\lfloor x\rfloor_+=\max(x,0)$. 
A positive score for a token therefore indicates that increasing the token's attention score is estimated to increase the target logit to first order. We collect these tokens in
$R_+^{\ell,h} = \{\, j \in \mathcal{I} : C_j^{\ell,h} > 0 \,\}$ and evaluate Eq.~\eqref{eq:wish-knockout} on it, giving \wish's head-ranking score:
\begin{equation}
s_t^{\ell,h}
=
\frac{\sum_{j\in\mathcal I}\lfloor C_j^{\ell,h}\rfloor_+}
     {1-\sum_{j\in R_+^{\ell,h}}A_j^{\ell,h}}
=
\frac{\sum_{j\in R_+^{\ell,h}} C_j^{\ell,h}}
     {1-\sum_{j\in R_+^{\ell,h}}A_j^{\ell,h}}.
\label{eq:wish-score}
\end{equation}
Higher scores indicate greater estimated decreases in the layer logits when the head's positive contribution is removed.
We discuss other interpretations for this score in  Supp.~\ref{app:knockout}.
\wish computes the scores for all heads and all layers resulting in $L\times H$ scores. \wish then ranks all heads of all layers in descending order of this score and selects specific heads based on this ordering. 

\noindent\textbf{Adaptive Head Selection.}
Having ranked the attention heads, we aim to select the highest-scoring heads for each target word. We adaptively determine how many of these heads to retain based on the distribution of their scores. 
The decreasing scores typically drop sharply among the top-ranked heads and flatten into a long tail of low-scoring heads (Fig.~\ref{fig:method-overview}(b)). We place the cutoff $i^\star$ at the knee of this curve, where the steep drop turns into the flat tail (Supp.~\ref{app:knee}), and retain the highest-ranked heads up to it:
\begin{equation}
S_t=\{(\ell,h):s_t^{\ell,h}>0,\;
\rho_t^{\ell,h}\leq i^\star\}.
\label{eq:wish-selected-heads}
\end{equation}
This determines the number of selected heads for each word independently, avoiding a tuned head budget as the best fixed budget varies across models (Supp.~\ref{app:budget}). We provide the implementation details for picking the knee in Supp.~\ref{app:implementation}.

\subsection{Spatial Map Construction}
\label{sec:aggregation}

We construct a heatmap from the selected head set $S_t$ for a target word $y_t$ through per-head spatial filtering of attention-gradient maps, followed by cross-head aggregation using rank-based weights.

\noindent\textbf{Per-Head Spatial Filtering.} Following DEX-AR~\cite{bousselham-dexar-2026}, we use positive attention gradients $P^{\ell,h}=\lfloor G^{\ell,h}\rfloor_+$ as the visual map for each selected head. 
Inspired by TAM's spatial denoising~\cite{li-token-activation-map-2025}, we apply a $3\times3$ median filter to each head's map suppressing isolated responses. For each visual token $j\in\mathcal I$, we apply the filtering on the map value $P_j^{\ell,h}$ as:
\begin{equation}
\widetilde P_j^{\ell,h} = \operatorname{median}\left(\{P_m^{\ell,h}\}_{m\in\mathcal N(j)}\right),
\label{eq:wish-head-map-filtered}
\end{equation}
where $\mathcal N(j)$ is the neighborhood of visual token $j$ on the image-token grid. The resulting filtered map is hence $\widetilde P^{\ell,h}$.

\noindent\textbf{Cross-Head Aggregation.} To aggregate the maps across the heads, we use a weighted median aggregation reducing sensitivity to unusually large responses from individual heads. We define weights that decrease linearly with the head rank $\rho_t^{\ell,h}$ as $\boldsymbol{\omega}_t$. 
The final heatmap $M_t$ is computed at each spatial position $j$ as: 
\begin{equation}
M_{t,j} = \operatorname{median}_{\boldsymbol{\omega}_t}\left(\{\widetilde P_j^{\ell,h}\}_{(\ell,h)\in S_t}\right).
\label{eq:wish-head-median}
\end{equation}
Finally, we min--max normalize $M_t$ over the visual-token positions to obtain the word-specific explanation. Implementation details are provided in  Supp.~\ref{app:implementation}.

\section{Function-Word Grounding}
\label{sec:funcwords}
\label{sec:fwp}
Function words that introduce nouns, such as ``a'' or ``the'', primarily serve grammatical roles. 
We investigate whether VLMs use object regions in an image when generating function words that precede nouns referring to objects in a caption. We call this ``look-ahead grounding'' and propose the following metrics to evaluate it.

\noindent\textbf{Function-Word Perturbation (FWP)}
We test whether the regions highlighted by a function word's map affect the noun it introduces more than the regions of other function words in the caption do. Following DEX-AR~\cite{bousselham-dexar-2026}, we obtain a perturbed image $I_b$ by replacing the pixels in a fraction $b$ of the image cells with the dataset mean pixel value.  We then measure the mean log-probability change of target word $y$ from the original image $I$ to the perturbed image $I_b$ on a set of token positions $\mathcal T$ as
\begin{equation}
\delta_{\mathcal T}(I_b)=\frac{1}{|\mathcal T|}\sum_{i\in\mathcal T}\log\frac{p_\theta(y_i\mid I,y_{<i})}{p_\theta(y_i\mid I_b,y_{<i})}.
\label{eq:log-ppl}
\end{equation}

Following this definition, consider a function word $y_t$ and the noun it introduces $y_n$. Let $\mathcal F$ be the set of positions of the other function words in the caption, each introducing a different object. For another function word $y_{t'}$ at a position $t'\in\mathcal F$, we obtain $I_b^t$ and $I_b^{t'}$ by replacing the fraction $b$ of cells ranked highest in the maps of $y_t$ and $y_{t'}$, respectively. We define FWP as the noun's change under its own function word's perturbed image minus its average change under those of the other function words:
\begin{equation}
\mathrm{FWP}(b)=\mathbb{E}_t\left[
\delta_{\{n\}}(I_b^t)
-\frac{1}{|\mathcal F|}
\sum_{t'\in\mathcal F}\delta_{\{n\}}(I_b^{t'})
\right].
\label{eq:fwp}
\end{equation}
The two terms differ only in which word's map selects the replaced cells, so a map that is the same for every word scores zero. Positive values indicate that a function word's map highlights evidence specific to the noun it introduces. We evaluate $b\in\{0,0.1,\dots,0.9\}$ and report the normalized area under the curve, $\mathrm{FWP\text{-}AUC}=\frac{1}{0.9}\int_0^{0.9}\mathrm{FWP}(b)\,db$, computed by trapezoidal integration. 

\noindent\textbf{Function-Word Segmentation (FWS)} 
\label{sec:fws}
To measure whether a function word's map highlights the object it introduces more than the other objects in the caption, we use Soft-IoU~\cite{bousselham-dexar-2026}. Given a function word $y_t$ and its associated object noun $y_n$, let the function word's explanation map $M_t$, the object's ground-truth segmentation mask by $\bar M_n$ and the masks of other objects introduced by function words $\mathcal O_t$. We define FWS as: 
\begin{equation}
\mathrm{FWS}_t=\text{Soft-IoU}(M_t,\bar M_n)
-\frac{1}{|\mathcal O_t|}
\sum_{\bar M\in\mathcal O_t}\text{Soft-IoU}(M_t,\bar M).
\label{eq:fws}
\end{equation}
Higher FWS indicates that the function word's map overlaps the introduced object more than the other objects named in the caption.
\section{Experiments}
\label{sec:experiments}

\begin{table*}[!t]
\centering\scriptsize
\resizebox{0.9\textwidth}{!}{
\renewcommand{\arraystretch}{1.08}
\setlength{\aboverulesep}{0.25ex}
\setlength{\belowrulesep}{0.4ex}
\captionsetup{skip=4pt}
\setlength{\tabcolsep}{0.8pt}
\begin{minipage}[t]{0.495\textwidth}
\centering
\resizebox{\linewidth}{!}{%
\begin{tabular}{@{}llcccccccc@{}}
\toprule
  &   & \multicolumn{2}{c}{ImageNet} & \multicolumn{2}{c}{VQAv2} & \multicolumn{2}{c}{COCO Caption} & \multicolumn{2}{c}{PASCAL VOC} \\
\cmidrule(lr){3-4}\cmidrule(lr){5-6}\cmidrule(lr){7-8}\cmidrule(lr){9-10}
 & Method & Del. $\uparrow$ & Ins.\ $\downarrow$ & Del. $\uparrow$ & Ins.\ $\downarrow$ & Del. $\uparrow$ & Ins.\ $\downarrow$ & Del. $\uparrow$ & Ins.\ $\downarrow$ \\
\midrule
\multirow{7}{*}{\rotatebox[origin=c]{90}{Qwen3-VL}} & Attention & 0.18 & 1.88 & \underline{0.80} & 0.53 & 1.02 & 1.41 & 1.19 & 1.05 \\
 & Attn$\times$Grad & 0.18 & 1.86 & 0.78 & 0.53 & 1.00 & 1.42 & 1.29 & 0.98 \\
 & Grad-CAM & 0.30 & 1.66 & 0.75 & 0.54 & \underline{1.10} & 1.27 & 1.24 & 0.94 \\
 & Chefer-CAM & 0.44 & 1.44 & 0.65 & 0.48 & 1.06 & 1.31 & 1.11 & 1.08 \\
 & DEX-AR & \underline{0.68} & 1.31 & 0.79 & 0.55 & \underline{1.10} & 1.25 & 1.54 & 0.95 \\
 & TAM & 0.62 & \underline{0.71} & 0.67 & \underline{0.45} & 1.01 & \underline{1.24} & \underline{1.67} & \underline{0.64} \\
\cmidrule(l){2-10}
 & \wish & \textbf{1.03} & \textbf{0.56} & \textbf{0.92} & \textbf{0.40} & \textbf{1.27} & \textbf{0.99} & \textbf{1.99} & \textbf{0.59} \\
\midrule
\multirow{7}{*}{\rotatebox[origin=c]{90}{LLaVA-1.5}} & Attention & 0.67 & 0.24 & 0.44 & 0.07 & 0.62 & 0.27 & 1.91 & 0.62 \\
 & Attn$\times$Grad & 0.71 & 0.19 & \underline{0.46} & \underline{0.06} & 0.64 & 0.26 & 2.20 & 0.53 \\
 & Grad-CAM & 0.68 & 0.20 & 0.45 & \textbf{0.05} & 0.62 & 0.26 & 1.77 & 0.60 \\
 & Chefer-CAM & 0.63 & 0.24 & \underline{0.46} & \textbf{0.05} & 0.61 & 0.26 & 1.76 & 0.59 \\
 & DEX-AR & 0.71 & 0.20 & 0.42 & 0.09 & \underline{0.65} & 0.29 & 2.48 & 0.48 \\
 & TAM & \underline{0.73} & \textbf{0.14} & 0.42 & 0.14 & \underline{0.65} & \underline{0.21} & \underline{2.66} & \underline{0.43} \\
\cmidrule(l){2-10}
 & \wish & \textbf{0.75} & \underline{0.16} & \textbf{0.49} & \textbf{0.05} & \textbf{0.72} & \textbf{0.19} & \textbf{2.71} & \textbf{0.37} \\
\midrule
\multirow{7}{*}{\rotatebox[origin=c]{90}{GLM-4.6V}} & Attention & 0.31 & 1.29 & 1.09 & 0.74 & 0.76 & 0.92 & 1.08 & 0.87 \\
 & Attn$\times$Grad & 0.32 & 1.27 & 1.14 & 0.70 & 0.76 & 0.92 & 1.16 & 0.85 \\
 & Grad-CAM & 0.40 & 1.19 & \textbf{1.32} & 0.67 & 0.88 & \underline{0.80} & \underline{1.21} & \underline{0.80} \\
 & Chefer-CAM & 0.37 & 1.10 & \underline{1.26} & \textbf{0.62} & 0.83 & 0.84 & 1.17 & 0.83 \\
 & DEX-AR & \textbf{0.82} & 0.74 & 1.24 & \underline{0.65} & \textbf{0.96} & \textbf{0.72} & 1.08 & 0.89 \\
 & TAM & 0.55 & \underline{0.60} & 0.90 & 0.86 & 0.68 & 0.95 & 1.03 & 0.92 \\
\cmidrule(l){2-10}
 & \wish & \underline{0.68} & \textbf{0.52} & 1.11 & 0.72 & \underline{0.93} & \textbf{0.72} & \textbf{1.32} & \textbf{0.62} \\
\bottomrule
\end{tabular}}
\end{minipage}
\hspace{4mm}
\begin{minipage}[t]{0.495\textwidth}
\centering
\resizebox{\linewidth}{!}{%
\begin{tabular}{@{}llcccccccc@{}}
\toprule
  &   & \multicolumn{2}{c}{ImageNet} & \multicolumn{2}{c}{VQAv2} & \multicolumn{2}{c}{COCO Caption} & \multicolumn{2}{c}{PASCAL VOC} \\
\cmidrule(lr){3-4}\cmidrule(lr){5-6}\cmidrule(lr){7-8}\cmidrule(lr){9-10}
 & Method & Del. $\uparrow$ & Ins.\ $\downarrow$ & Del. $\uparrow$ & Ins.\ $\downarrow$ & Del. $\uparrow$ & Ins.\ $\downarrow$ & Del. $\uparrow$ & Ins.\ $\downarrow$ \\
\midrule
\multirow{7}{*}{\rotatebox[origin=c]{90}{BakLLaVA}} & Attention & 2.12 & 1.50 & 0.42 & 0.04 & 1.12 & 0.45 & 2.63 & 0.73 \\
 & Attn$\times$Grad & 2.22 & 1.43 & 0.43 & 0.03 & \underline{1.14} & 0.44 & 2.85 & 0.63 \\
 & Grad-CAM & \textbf{2.25} & 1.43 & 0.43 & 0.04 & 1.13 & 0.43 & 2.57 & 0.71 \\
 & Chefer-CAM & \underline{2.24} & 1.39 & 0.45 & \underline{0.02} & \underline{1.14} & 0.44 & 2.56 & 0.70 \\
 & DEX-AR & 2.23 & 1.28 & 0.37 & 0.06 & 0.96 & 0.57 & \underline{3.05} & 0.59 \\
 & TAM & 1.87 & \underline{1.16} & \underline{0.47} & \underline{0.02} & 1.02 & \underline{0.42} & 3.01 & \underline{0.56} \\
\cmidrule(l){2-10}
 & \wish & 2.14 & \textbf{1.06} & \textbf{0.49} & \textbf{0.01} & \textbf{1.20} & \textbf{0.34} & \textbf{3.36} & \textbf{0.48} \\
\midrule
\multirow{7}{*}{\rotatebox[origin=c]{90}{OV-1.5}} & Attention & 1.62 & 1.99 & 0.29 & $-$0.01 & \underline{0.72} & 0.45 & 1.51 & 0.73 \\
 & Attn$\times$Grad & 1.73 & 1.91 & \underline{0.30} & $-$0.01 & \underline{0.72} & 0.45 & 1.64 & 0.72 \\
 & Grad-CAM & 1.79 & 1.90 & 0.27 & \underline{$-$0.02} & 0.70 & 0.47 & 1.38 & 0.77 \\
 & Chefer-CAM & 1.51 & 1.97 & 0.23 & 0.01 & \underline{0.72} & 0.45 & 1.37 & 0.79 \\
 & DEX-AR & \textbf{2.04} & 1.68 & 0.19 & 0.07 & 0.65 & 0.53 & \underline{1.69} & 0.86 \\
 & TAM & 1.72 & \underline{1.19} & 0.22 & 0.08 & 0.68 & \underline{0.40} & 1.51 & \underline{0.55} \\
\cmidrule(l){2-10}
 & \wish & \underline{1.96} & \textbf{1.05} & \textbf{0.39} & \textbf{$-$0.03} & \textbf{0.77} & \textbf{0.35} & \textbf{1.98} & \textbf{0.49} \\
\midrule
\multirow{7}{*}{\rotatebox[origin=c]{90}{OV-2}} & Attention & 0.70 & 1.00 & \underline{0.71} & 0.27 & \underline{0.63} & 0.40 & 1.32 & 0.72 \\
 & Attn$\times$Grad & 0.75 & 0.94 & \textbf{0.72} & \underline{0.25} & 0.61 & 0.41 & 1.46 & 0.68 \\
 & Grad-CAM & 0.74 & 0.98 & 0.70 & 0.28 & 0.56 & 0.47 & 1.06 & 0.82 \\
 & Chefer-CAM & 0.73 & 0.97 & 0.66 & 0.26 & 0.62 & 0.40 & 1.16 & 0.73 \\
 & DEX-AR & 0.97 & 0.76 & 0.53 & 0.36 & 0.50 & 0.50 & 1.50 & 0.78 \\
 & TAM & \underline{1.13} & \underline{0.57} & 0.58 & 0.33 & 0.59 & \underline{0.38} & \underline{1.61} & \underline{0.53} \\
\cmidrule(l){2-10}
 & \wish & \textbf{1.25} & \textbf{0.51} & 0.70 & \textbf{0.24} & \textbf{0.72} & \textbf{0.32} & \textbf{1.91} & \textbf{0.51} \\
\bottomrule
\end{tabular}}
\end{minipage}
}
\caption{Image perturbation evaluation across six VLMs on ImageNet, VQAv2, COCO Caption and PASCAL VOC. Values are the area under the log normalized-perplexity curve for deletion of the most relevant pixels first (Del., $\uparrow$) and insertion into a blurred image (Ins., $\downarrow$). Best value per model in bold, second best underlined.}
\label{tab:perturbation}
\vspace{-5mm}
\end{table*}

\subsection{Experimental Setup}
\label{sec:tasks}

\noindent\textbf{Protocol.} We evaluate word-level explanations in free generation and teacher forced settings across different tasks. For image captioning, each model generates a one-sentence description using greedy decoding. For classification and visual question answering, we teacher force the  class labels or the annotated answers. For evaluations requiring object annotations, we identify the object words by matching their lemmas to annotated object classes with segmentation masks in the image. All methods explain the same targets at their respective generation positions. Further details are in Supp.~\ref{app:targets}.

\noindent\textbf{Datasets.} We evaluate on four datasets. PASCAL VOC 2012~\cite{everingham-voc-2010} provides 1{,}449 validation images with object annotations for 20 classes, which we use for evaluating head selection, localization, perturbation, and function-word grounding. For COCO Captions~\cite{chen-coco-captions-2015}, we adopt TAM's 5{,}000-image evaluation split~\cite{li-token-activation-map-2025} with COCO object annotations~\cite{lin-coco-2014}, using all images for localization and a randomly selected 1{,}000 subset for perturbation and head selection. Finally, following DEX-AR~\cite{bousselham-dexar-2026}, we use 5{,}000 ImageNet~\cite{deng-imagenet-2009} validation images and 1{,}000 VQAv2~\cite{goyal-vqav2-2017} validation image--question pairs for perturbation evaluation.

\noindent\textbf{Models.} To assess \wish across diverse visual encoders and language-model backbones, we evaluate six VLMs: LLaVA-1.5-7B~\cite{liu-llava15-2024}, BakLLaVA~\cite{skunkworks-bakllava-2023}, Qwen3-VL-8B~\cite{qwen3-vl-2025}, GLM-4.6V-Flash~\cite{vteam-glm45v-2025}, LLaVA-OneVision-1.5-8B~\cite{an-llava-onevision-15-2025}, and LLaVA-OneVision-2-8B~\cite{an-llava-onevision-2-2026}.

\subsection{Comparison to State-of-the-Art}
\label{sec:explanation_eval}

We compare the quality of \wish's spatial explanations for the case of predictive faithfulness measured through image perturbation tests, and spatial localization measured through segmentation-based tests. 
We compare against TAM~\cite{li-token-activation-map-2025} and DEX-AR~\cite{bousselham-dexar-2026}, using their released implementations, and our implementations of Attention, Attn$\times$Grad~\cite{chefer-beyond-attention-2021}, Grad-CAM~\cite{selvaraju-gradcam-2017}, and Chefer-CAM~\cite{chefer-generic-attention-2021} (Supp.~\ref{supp:baselines}).
We also report \wish with fixed head budgets in the ablations (Sec.~\ref{sec:ablations}) and in Supp.~\ref{app:budget}.

\begin{table}[!t]
\centering\scriptsize
\renewcommand{\arraystretch}{0.96}
\setlength{\aboverulesep}{0.25ex}
\setlength{\belowrulesep}{0.4ex}
\captionsetup{skip=4pt}
\setlength{\tabcolsep}{1.8pt}
\begin{tabular*}{\columnwidth}{@{\extracolsep{\fill}}clcccccc@{}}
\toprule
  &   & \multicolumn{3}{c}{PASCAL VOC} & \multicolumn{3}{c}{COCO Caption} \\
\cmidrule(lr){3-5}\cmidrule(lr){6-8}
 & Method & Obj-IoU & Soft-IoU & EPG & Obj-IoU & Soft-IoU & EPG \\
\midrule
\multirow{7}{*}{\rotatebox[origin=c]{90}{Qwen3-VL}} & Attention & 14.2 & 8.6 & 30.2 & 10.4 & 6.1 & 22.0 \\
 & Attn$\times$Grad & 8.6 & 6.0 & 29.7 & 7.0 & 4.6 & 23.6 \\
 & Grad-CAM & 12.8 & 8.1 & 28.1 & 9.3 & 5.6 & 19.9 \\
 & Chefer-CAM & 6.6 & 5.5 & 24.8 & 4.7 & 3.7 & 17.6 \\
 & DEX-AR & 14.0 & 9.9 & 26.0 & 12.7 & 7.8 & 18.8 \\
 & TAM & \underline{21.8} & \underline{14.6} & \textbf{33.5} & \underline{18.9} & \underline{11.9} & \textbf{26.2} \\
\cmidrule(l){2-8}
 & \wish & \textbf{30.0} & \textbf{19.9} & \underline{32.7} & \textbf{24.6} & \textbf{15.6} & \underline{25.2} \\
\midrule
\multirow{7}{*}{\rotatebox[origin=c]{90}{LLaVA-1.5}} & Attention & 11.2 & 5.4 & 38.1 & 12.0 & 5.6 & 27.5 \\
 & Attn$\times$Grad & 20.0 & 9.7 & \underline{52.2} & 20.5 & 9.6 & \underline{38.1} \\
 & Grad-CAM & 17.3 & 8.0 & 37.8 & 18.2 & 7.7 & 26.7 \\
 & Chefer-CAM & 18.3 & 8.6 & 41.1 & 20.2 & 8.9 & 31.3 \\
 & DEX-AR & 40.5 & 20.0 & 31.5 & 28.9 & 14.4 & 20.0 \\
 & TAM & \underline{49.8} & \textbf{31.2} & \textbf{57.2} & \textbf{35.7} & \textbf{22.0} & \textbf{39.4} \\
\cmidrule(l){2-8}
 & \wish & \textbf{50.9} & \underline{28.2} & 51.0 & \underline{35.0} & \underline{19.8} & 33.0 \\
\midrule
\multirow{7}{*}{\rotatebox[origin=c]{90}{GLM-4.6V}} & Attention & 13.4 & 8.0 & \textbf{29.7} & 12.4 & 6.7 & 22.3 \\
 & Attn$\times$Grad & 7.3 & 5.2 & \underline{29.5} & 8.4 & 5.2 & \textbf{24.8} \\
 & Grad-CAM & 14.6 & 8.6 & 28.6 & 13.2 & 6.9 & 21.1 \\
 & Chefer-CAM & 10.5 & 6.7 & \textbf{29.7} & 10.6 & 5.9 & \underline{22.7} \\
 & DEX-AR & 8.6 & 8.3 & 23.3 & 11.3 & 7.9 & 16.6 \\
 & TAM & \underline{14.9} & \underline{10.7} & 26.6 & \underline{15.0} & \underline{9.5} & 21.4 \\
\cmidrule(l){2-8}
 & \wish & \textbf{22.9} & \textbf{14.7} & 29.4 & \textbf{18.8} & \textbf{11.7} & 22.6 \\
\midrule
\multirow{7}{*}{\rotatebox[origin=c]{90}{BakLLaVA}} & Attention & 21.7 & 8.8 & 39.7 & 20.5 & 8.1 & 27.9 \\
 & Attn$\times$Grad & 27.4 & 12.2 & \underline{51.4} & 25.5 & 11.1 & \underline{37.6} \\
 & Grad-CAM & 25.2 & 10.9 & 41.8 & 22.8 & 9.6 & 28.3 \\
 & Chefer-CAM & 21.7 & 9.7 & 41.8 & 21.7 & 9.1 & 29.7 \\
 & DEX-AR & 40.7 & 20.8 & 34.5 & 30.8 & 15.3 & 23.7 \\
 & TAM & \underline{43.4} & \underline{28.2} & \textbf{54.8} & \underline{35.2} & \textbf{23.4} & \textbf{41.1} \\
\cmidrule(l){2-8}
 & \wish & \textbf{50.1} & \textbf{28.7} & 50.4 & \textbf{37.0} & \underline{21.6} & 36.4 \\
\midrule
\multirow{7}{*}{\rotatebox[origin=c]{90}{OV-1.5}} & Attention & \underline{22.9} & 12.2 & 33.8 & 18.0 & 8.3 & 24.5 \\
 & Attn$\times$Grad & 16.1 & 9.3 & \textbf{35.0} & 16.4 & 8.4 & \underline{29.2} \\
 & Grad-CAM & 17.8 & 9.8 & 30.3 & 14.0 & 7.2 & 21.1 \\
 & Chefer-CAM & 15.2 & 9.0 & 29.3 & 14.7 & 7.5 & 22.4 \\
 & DEX-AR & 18.7 & 11.7 & 27.3 & 17.5 & 9.3 & 20.6 \\
 & TAM & 21.4 & \underline{14.5} & 33.8 & \underline{23.3} & \underline{14.4} & \textbf{30.4} \\
\cmidrule(l){2-8}
 & \wish & \textbf{33.2} & \textbf{20.4} & \underline{34.0} & \textbf{27.5} & \textbf{16.1} & 26.9 \\
\midrule
\multirow{7}{*}{\rotatebox[origin=c]{90}{OV-2}} & Attention & 12.3 & 8.0 & 31.7 & 5.6 & 4.3 & 22.7 \\
 & Attn$\times$Grad & 12.8 & 8.0 & \textbf{35.4} & 13.8 & 7.5 & \textbf{29.3} \\
 & Grad-CAM & 11.8 & 7.5 & 27.3 & 10.0 & 5.7 & 19.6 \\
 & Chefer-CAM & 11.6 & 7.7 & 28.3 & 11.3 & 6.3 & 21.5 \\
 & DEX-AR & 18.2 & 11.3 & 27.3 & 17.4 & 9.6 & 20.2 \\
 & TAM & \underline{23.2} & \underline{15.1} & 33.5 & \underline{23.3} & \underline{14.1} & \underline{28.4} \\
\cmidrule(l){2-8}
 & \wish & \textbf{32.4} & \textbf{20.4} & \underline{33.8} & \textbf{26.2} & \textbf{15.5} & 26.4 \\
\bottomrule
\end{tabular*}
\caption{Localization performance comparison on PASCAL VOC and COCO Caption for the object words of each model's own caption. Reported metrics are Obj-IoU ~\cite{li-token-activation-map-2025}, Soft-IoU and EPG~\cite{bousselham-dexar-2026}.}
\label{tab:localization}
\vspace{-5mm}
\end{table}

\subsubsection{Image Perturbation}
\label{sec:explanation_eval_perturbation}
We first evaluate whether explanation maps identify image regions that influence the target word prediction through image perturbation tests in Tab.~\ref{tab:perturbation}. 
\noindent\textbf{Metric.} We use DEX-AR's positive deletion (Del.) and insertion (Ins.) tests~\cite{bousselham-dexar-2026}. Deletion replaces pixels ranked highest by the map with the dataset mean pixel value while insertion restores the fraction of highest ranked pixels from a gaussian blurred image. For each perturbation fraction
$b \in \{0, 0.1, \dots, 0.9\}$, we compute $\delta_{\mathcal T}(I_b)$
(Eq.~\eqref{eq:log-ppl}) over all target positions $\mathcal T$, i.e., the
log of the perplexity ratio between the perturbed and the original image.
We report the area under $\delta_{\mathcal T}(I_b)$ as a function of $b$. Higher is better for deletion, lower for insertion.
\noindent\textbf{Results.}
\wish beats both TAM and DEX-AR in 40 of 48 comparisons. The exceptions cluster in two places. On ImageNet, DEX-AR deletes more effectively on three models while TAM inserts better on LLaVA-1.5. On GLM-4.6V, DEX-AR leads on both VQAv2 metrics and COCO deletion but ties on COCO insertion. \wish's advantage is thus largest for explaining individual words in freely-generated captions and smallest for teacher forced single class names on object-centric images. 

\subsubsection{Object Segmentation}
\label{sec:explanation_eval_segmentation}
\begin{figure}[t]
\centering
\includegraphics[width=\linewidth]{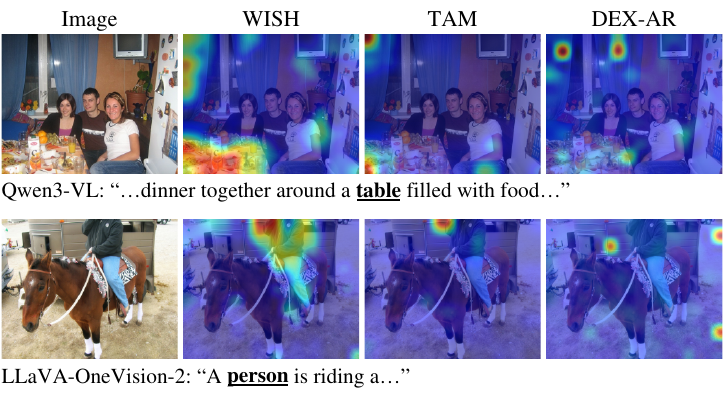}
\vspace{-8mm}
\caption{Visual explanations of object words in the models' generated captions comparing \wish, TAM, and DEX-AR. Explained words are underlined and bold.}
\label{fig:qual}
\vspace{-5mm}
\end{figure}

Second, we test whether the map of an object word covers the object it names in Tab.~\ref{tab:localization}. \noindent\textbf{Metric.}
For each object word of the model's own caption, we compare its map with the annotated mask of the object through three metrics. Obj-IoU follows TAM's protocol~\cite{li-token-activation-map-2025} and uses the best sub-token's binarized map, while Soft-IoU and Energy Pointing Game (EPG) follow DEX-AR~\cite{bousselham-dexar-2026} measuring continuous map overlap and energy concentration within the object mask respectively. 
\noindent\textbf{Results.}
\wish's maps overlap the named object best in most settings. On PASCAL VOC, it outperforms every baseline in Obj-IoU on all six models by 1.1--10.3 points. On COCO, it does so on five of six models. It also achieves a higher Soft-IoU than every baseline in 9 of 12 comparisons. However, \wish doesn't lead on EPG suggesting that its maps cover the object more completely at the cost of some energy outside it. Qualitative examples of object words are shown in Fig.~\ref{fig:qual}.

\subsection{Head Selection Evaluation}
\label{sec:head_selection}

\begin{figure}[t]
  \centering
  \includegraphics[width=\linewidth]{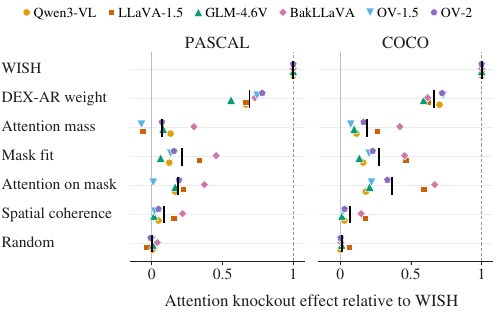}
\vspace{-6mm}
\caption{Image-attention knockout across head-selection criteria on PASCAL and COCO object words. We report the drop in target-word log-probability when blocking attention to the whole image, relative to \wish. All head sets are of equal size. Markers show individual models and black bars the six-model means. \wish's effect exceeds every other criterion on every model.}
  \label{fig:knockout}
\vspace{-5mm}
\end{figure}

To evaluate the effect of the heads selected by \wish on word predictions, we apply attention knockout~\cite{kaduri-whats-in-the-image-2025}, blocking the attention of the selected heads $S_t$ to the entire image tokens ($R=\mathcal I$) and measuring the resulting drop in the target word's final log-probability.  
 \noindent\textbf{Metric.} For a target word $y_t$,  we define the drop in final log-probability after the knockout as $\delta_t(S_t,R) = \log p_\theta(y_t\mid I,y_{<t}) - \log p_\theta^{\setminus(S_t,R)}(y_t\mid I,y_{<t})$, where $p_\theta^{\setminus(S_t,R)}$ denotes the probability after blocking attention to region $R$ in the selected heads at the predicting position. Here, $p_\theta^{\setminus(S_t,R)}$ is the model's output probability after jointly blocking attention in every head in $S_t$. This metric measures the downstream effect of the selected head set.
\noindent\textbf{Baselines.}
We compare \wish's head selection criterion against five alternative criteria and random selection, each selecting as many heads as \wish. We adapt \emph{DEX-AR weight} that ranks heads by the relative magnitude of image and text gradients~\cite{bousselham-dexar-2026}, \emph{attention mass} by their total visual attention and \emph{spatial coherence} by their gradient-map autocorrelation using Moran's $I$~\cite{moran1950notes}. We also test two ways to rank heads using ground-truth object masks. \emph{Mask fit} measures gradient energy inside the mask relative to its area, and \emph{attention on mask} measures how much attention
falls inside the mask. Exact criteria 
definitions are provided in Supp.~\ref{app:head-rankings}. 
\noindent\textbf{Results.}
Blocking the heads selected by \wish from attending to the whole image lowers the target word's log-probability by 0.21--0.88 nats on PASCAL and 0.53--1.75 nats on COCO, whereas random sets of the same size change it by at most 0.04 nats. On both datasets, \wish exceeds every other criterion on all six models (Fig.~\ref{fig:knockout}). Averaged across models, DEX-AR's weighting comes closest at 69\% (PASCAL) and 66\% (COCO) of \wish's effect, while the remaining criteria, including the two that use the ground-truth mask, reach at most 37\%. These results suggest that attending strongly to the image or producing a well-localized map does not necessarily make a head influential for the target word.

\subsection{Function-Word Grounding Evaluation}
\label{sec:eval-funcwords}

\begin{figure}[!t]
\centering
\includegraphics[width=\linewidth]{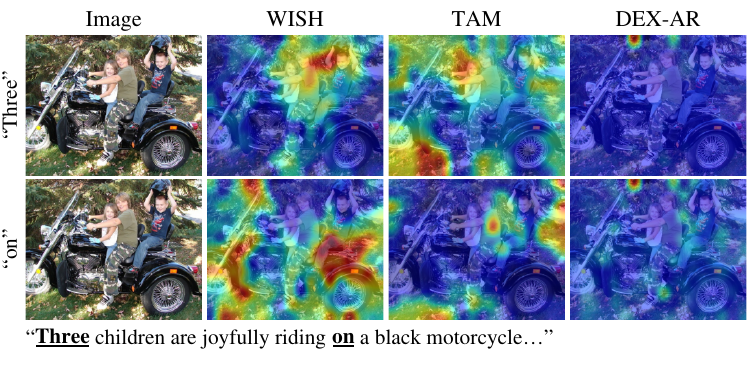}
\vspace{-8mm}
\caption{Function words of different categories in a LLaVA-OneVision-1.5 caption: the numeral ``Three'' introduces \emph{children} and the preposition ``on'' introduces \emph{motorcycle}, each explained before the noun it introduces is generated.}
\label{fig:det}
\vspace{-4mm}
\end{figure}

We use Stanza~\cite{qi-stanza-2020} to parse function words and their associated nouns in model-generated captions through Universal Dependencies (UD)~\cite{nivre-ud-2020}. More specifically, we focus on function words that attach to a noun naming an annotated object.  These include determiners (\texttt{det}), prepositions (\texttt{case}), coordinating conjunctions (\texttt{cc}), and numerals (\texttt{nummod}). 
We further require the word to precede its noun and the noun to be the first mention of its object class, so that each word introduces an object not yet named in the caption and its explanation is computed before that object is named. We keep the first function word that introduces each object and evaluate captions in which at least two objects are introduced this way. Further details on function-word selection and excluded categories are provided in  Supp.~\ref{app:function-words}.
We report FWP-AUC and FWS (Sec.~\ref{sec:fws}) and positive deletion (Sec.~\ref{sec:explanation_eval_perturbation}) with the function word as the target, $\mathcal T=\{t\}$, which measures whether the regions highlighted for a function word affect its own prediction. All reported results are averaged over the six models. 
\noindent\textbf{Results.} Tab.~\ref{tab:function_words} shows that FWP-AUC is positive for every method. 
Thus, removing the region highlighted for a function word lowers the probability of the noun it introduces more than removing the regions highlighted for the other function words. 
\wish captures this best by scoring 0.52 against 0.35 for Grad-CAM (difference 0.17 [0.06, 0.27]). The same regions also matter for the function word itself, where \wish again leads (Del.\ 1.57 against 1.36 for Grad-CAM). On FWS, \wish scores 2.68 against 1.22 for Attn$\times$Grad (difference 1.47 [0.92, 2.00]). \wish leads on five of six models on each metric, while TAM and DEX-AR score lowest on FWP-AUC, as their function-word maps change little between words. Together, these results indicate that VLMs already rely on the region of the upcoming object when generating the function word that introduces it. Fig.~\ref{fig:det} shows qualitative examples from different categories of function-words (per-model 
results in Supp.~\ref{app:function-words}.).
\begin{table}[t]
  \centering
  \footnotesize
  \setlength{\tabcolsep}{3pt}
  \begin{tabular}{@{}lccc@{}}
    \toprule
    Method & Del.$\uparrow$ & FWP-AUC$\uparrow$ & FWS$\uparrow$ \\
    \midrule
    Attention & 1.24 [1.16, 1.32] & 0.34 [0.28, 0.40] & 0.70 [0.48, 0.95] \\
    Attn$\times$Grad & 1.26 [1.17, 1.35] & \underline{0.35} [0.29, 0.41] & \underline{1.22} [1.02, 1.41] \\
    Grad-CAM & \underline{1.36} [1.26, 1.47] & \underline{0.35} [0.30, 0.40] & 0.88 [0.67, 1.10] \\
    Chefer-CAM & 1.33 [1.24, 1.43] & 0.32 [0.26, 0.37] & 0.91 [0.67, 1.16] \\
    DEX-AR & 1.32 [1.23, 1.42] & 0.12 [0.04, 0.20] & 0.77 [0.25, 1.26] \\
    TAM & 1.03 [0.93, 1.12] & 0.16 [0.06, 0.26] & 1.05 [0.43, 1.68] \\
    \midrule
    \wish & \textbf{1.57} [1.46, 1.69] & \textbf{0.52} [0.41, 0.62] & \textbf{2.68} [2.10, 3.24] \\
    \bottomrule
  \end{tabular}
  \vspace{-2mm}
  \caption{Function-word grounding on PASCAL VOC on 2{,}984 nouns for captions with at least two objects. Positive deletion with the function word as target (Del.), FWP-AUC (Eq.~\eqref{eq:fwp}) and FWS (Eq.~\eqref{eq:fws}) in percentages. Averaged over function-word categories and six models; 95\% bootstrap intervals in square brackets. }
  \label{tab:function_words}
  \vspace{-4mm}
\end{table}

\subsection{Ablation Studies}
\label{sec:ablations}

We ablate the contribution of \wish's head-selection, spatial filtering and aggregation components by changing one component at a time while keeping the others fixed. Tab.~\ref{tab:ablations} reports results on PASCAL VOC, averaged across six models.  The best fixed budget $K=64$ was tuned based on PASCAL VOC training set (Supp.~\ref{app:budget}). \noindent\textbf{Head selection.} The adaptively selected set based on the knee matches the best fixed budget of 64 heads, achieving 2.21 positive-deletion AUC and 36.6 Obj-IoU compared with 2.21 and 36.7 for Top-64.  This shows that the knee comes close to the best tuned budget without tuning, since it adapts the number of heads to each word. Selecting all positive-scoring heads reduces these values to 1.76 and 26.5, while random selection at the knee's budget yields 1.05 and 17.4. Thus, maps built from the adaptively selected heads are better on all metrics than not selecting at all (all heads). We further compare the score (Eq.~\eqref{eq:wish-score}) with the exact attention knockout score it approximates (Eq.~\eqref{eq:attention-knockout-score}).  The Spearman correlation between the scores and measured drops is at least $0.999$ in median across target words
for each of the six models (Supp.~\ref{supp:baselines}ig.~\ref{fig:eq4-exact}). \noindent\textbf{Spatial Filtering.} We assess whether filtering noisy, isolated visual tokens in each head's gradient map improves explanation quality. Removing the $3\times3$ median filter reduces positive-deletion AUC by 0.18 and Obj-IoU by 7.9 points. Replacing it with mean or Gaussian filtering also lowers all four metrics, suggesting that the median filter is more effective at suppressing outliers while preserving spatial structure. \noindent\textbf{Aggregation.}  Comparing different aggregation methods yields similar results. Across the rank-weighted median, rank-weighted mean, unweighted median, and mean, positive-deletion AUC varies by at most 0.05 and Obj-IoU by 0.8 points. However, the rank-weighted median achieves the highest EPG (38.5), compared with 35.3, 37.2, and 34.2, respectively. The per-model results are provided in the supplement (Supp.~\ref{app:additional}).
\begin{table}[t]
\centering
\setlength{\tabcolsep}{2.2pt}
\footnotesize
\begin{tabular}{@{}clcccc@{}}
\toprule
 & Variant & Del.$\uparrow$ & Obj-IoU$\uparrow$ & Soft-IoU$\uparrow$ & EPG$\uparrow$ \\
\midrule
\multirow{4}{*}{\rotatebox[origin=c]{90}{\scriptsize Selection}}
 & Random (knee-sized) & 1.05 & 17.4 & 12.5 & 25.3 \\
 & All heads & \underline{1.76} & 26.5 & 18.8 & 27.6 \\
 & $s_t^{\ell,h}$: Top-64 & \textbf{2.21} & \textbf{36.7} & \textbf{22.1} & \underline{38.4} \\
 & \textit{(Ours)} $s_t^{\ell,h}$: Knee & \textbf{2.21} & \underline{36.6} & \underline{22.0} & \textbf{38.5} \\
\midrule
\multirow{4}{*}{\rotatebox[origin=c]{90}{\scriptsize Filtering}}
 & None & 2.03 & 28.7 & 16.1 & \underline{36.2} \\
 & $3{\times}3$ mean & 2.12 & \underline{33.5} & \underline{20.6} & 31.1 \\
 & $3{\times}3$ Gaussian ($\sigma=1$) & \underline{2.13} & \underline{33.5} & 20.3 & 31.3 \\
 & \textit{(Ours)} $3{\times}3$ median & \textbf{2.21} & \textbf{36.6} & \textbf{22.0} & \textbf{38.5} \\
\midrule
\multirow{4}{*}{\rotatebox[origin=c]{90}{\scriptsize Aggregation}}
 & Mean & 2.20 & 35.8 & 20.0 & 34.2 \\
 & Median & 2.18 & \underline{36.4} & \textbf{22.3} & \underline{37.2} \\
 & Rank-weighted mean & \textbf{2.23} & 35.8 & 20.2 & 35.3 \\
 & \textit{(Ours)} Rank-weighted median & \underline{2.21} & \textbf{36.6} & \underline{22.0} & \textbf{38.5} \\
\bottomrule
\end{tabular}
\vspace{-2mm}
\caption{Controlled ablation of \wish on PASCAL VOC (six-model means). 
Heads are ranked by the \wish score $s_t^{\ell,h}$, where \emph{Knee} is \wish's adaptive head selection and \emph{Top-64} keeps a fixed budget of the 64 highest-scoring heads. \textit{(Ours)} marks \wish's default choice, shown last in each block.}
\label{tab:ablations}
\vspace{-6mm}
\end{table}
\section{Conclusion}\label{sec:conclusion}
We propose \wish as a method for head-selection based explainability for VLMs. Our results show that head selection improves explainability and that  a head's importance can be estimated from its contribution to the target word. This allows us to show that, similar to LLMs, VLMs also exhibit look-ahead grounding, indicating that VLMs already rely on an object's image region while generating the function word that introduces it, before naming the object.

\section*{LLM Usage Statement}
We used OpenAI's ChatGPT and Anthropic's Claude to edit and refine author-produced manuscript text, implementation
code and figures and diagrams. We did not use LLMs or other
generative AI tools to design or run experiments, analyze results,
create or annotate datasets, or develop research ideas, hypotheses,
claims, or conclusions. We did use a locally hosted Qwen3.6-27B and shared major portions of the manuscript to it.  
\newpage
{
    \small
    \bibliographystyle{ieeenat_fullname}
    \bibliography{references}
}
\clearpage
\appendix

\setcounter{figure}{0}\setcounter{table}{0}\setcounter{equation}{0}
\renewcommand{\thefigure}{S\arabic{figure}}
\renewcommand{\thetable}{S\arabic{table}}
\renewcommand{\theequation}{S\arabic{equation}}

\renewcommand{\dblfloatpagefraction}{0.5}
\makeatletter
\renewcommand\paragraph{\@startsection{paragraph}{4}{\z@}{0.6ex \@plus .3ex \@minus .3ex}{-0.6em}{\normalfont\normalsize\bfseries}}
\setlength{\@fptop}{0pt}
\setlength{\@dblfptop}{0pt}
\makeatother
\exhyphenpenalty=10000
\widowpenalty=10000 \clubpenalty=10000
\hyphenation{pascal eagle llava bakllava}
\AddToHook{env/figure/begin}{\hyphenpenalty=10000 }
\AddToHook{env/figure*/begin}{\hyphenpenalty=10000 }
\AddToHook{env/table/begin}{\hyphenpenalty=10000 }
\AddToHook{env/table*/begin}{\hyphenpenalty=10000 }

\makeatletter
\let\wishorig@addcontentsline\addcontentsline
\renewcommand\addcontentsline[3]{%
  \wishorig@addcontentsline{#1}{#2}{#3}%
  \def\wish@tmpa{toc}\def\wish@tmpb{#1}%
  \ifx\wish@tmpa\wish@tmpb
    \begingroup\let\texorpdfstring\@secondoftwo%
    \addtocontents{stoc}{\protect\contentsline{#2}{#3}{\thepage}{\@currentHref}}%
    \endgroup
  \fi}
\twocolumn[{%
\centering
{\Large\bfseries WISH: Word-Informed Selection of Heads for Vision-Language Explainability\par}
\vspace{0.5em}
{\Large Supplementary Material\par}
\vspace{1.5em}
\begin{minipage}{\textwidth}
\raggedright
{\large\bfseries Contents\par}
\vspace{0.8em}
\fontsize{10}{14.5}\selectfont
\setlength{\parskip}{0pt}
\renewcommand*\numberline[1]{\hb@xt@\@tempdima{#1.\hfil}}
\renewcommand*\l@section[2]{\addvspace{0.45em}\@dottedtocline{1}{0em}{1.5em}{\bfseries #1}{#2}}
\renewcommand*\l@subsection[2]{\@dottedtocline{2}{1.5em}{2.3em}{#1}{#2}}
\@starttoc{stoc}
\end{minipage}
}]
\makeatother
\clearpage

\section{Notation}
\label{app:notation}

Tab.~\ref{tab:app-notation} lists the symbols used in the main paper and in this supplementary material.
Citation numbers refer to the reference list of the main paper.

\begin{table*}[p]
\centering
\footnotesize
\renewcommand{\arraystretch}{1.15}
\begin{tabular}{@{}lp{0.70\textwidth}@{}}
\toprule
Symbol & Meaning \\
\midrule
\multicolumn{2}{@{}l}{\emph{Model and attention (Sec.~\ref{sec:selection} of the main paper)}} \\
$L$, $H$ & number of layers and number of attention heads per layer \\
$\ell$, $h$ & layer index and head index \\
$t$, $y_t$, $y_{<t}$ & generation step, the token generated at step $t$ (the target word) and the tokens before it \\
$\mathbf z_i^\ell\in\mathbb R^d$ & token representation at sequence position $i$ after layer $\ell$, with dimension $d$ \\
$\mathbf Q^{\ell,h}$, $\mathbf K^{\ell,h}$, $d_k$ & query and key projections of head $(\ell,h)$, with key dimension $d_k$ per head \\
$\mathbf e^{\ell,h}$, $\mathbf A^{\ell,h}$ & pre-softmax attention scores and attention weights of head $(\ell,h)$ \\
$A_{i,j}^{\ell,h}$ & attention weight from query position $i$ to key position $j$ \\
$e_k^{\ell,h}$, $A_k^{\ell,h}$ & attention scores and attention weights at the predicting row $i=-1$, to key $k$ \\
$\mathcal K$ & key positions accessible from the predicting row including visual and text tokens \\
$\mathcal I\subset\mathcal K$ & positions of the visual tokens $j$. The text tokens are $\mathcal K\setminus\mathcal I$ \\
$\mathbf W_{\mathrm U}$, $\mathcal V$ & unembedding matrix and vocabulary \\
$\operatorname{Norm}$ & the model's final normalization \\
$\hat o^{\ell}(y_t)$ & layer logit (logit lens) of $y_t$ at layer $\ell$ \\
$p_\theta(y_t\mid I,y_{<t})$ & the model's output probability of $y_t$ given the image $I$ and the preceding tokens \\
\midrule
\multicolumn{2}{@{}l}{\emph{Head-ranking score (Sec.~\ref{sec:selection} of the main paper, Sec.~\ref{app:knockout})}} \\
$R\subset\mathcal I$ & a set of visual tokens removed from one head's attention \\
$\hat o^{\ell,\setminus(h,R)}(y_t)$ & layer logit after blocking head $(\ell,h)$'s attention to $R$ \\
$\Delta^{\ell,h}(R)$ & attention knockout score $\hat o^{\ell}(y_t)-\hat o^{\ell,\setminus(h,R)}(y_t)$ \\
$G_k^{\ell,h}$ & attention-gradient $\partial\hat o^{\ell}(y_t)/\partial A_k^{\ell,h}$, $k\in\mathcal K$ \\
$\bar G^{\ell,h}$ & attention-weighted mean gradient over all keys, $\sum_{k\in\mathcal K}A_k^{\ell,h}G_k^{\ell,h}$ \\
$C_j^{\ell,h}$ & token's score $A_j^{\ell,h}(G_j^{\ell,h}-\bar G^{\ell,h})=\partial\hat o^{\ell}(y_t)/\partial e_j^{\ell,h}$ \\
$R_+^{\ell,h}$ & visual tokens with positive scores $\{j\in\mathcal I:C_j^{\ell,h}>0\}$ \\
$s_t^{\ell,h}$ & \wish head-ranking score \\
$\lfloor x\rfloor_+$ & $\max(x,0)$ / ReLU(x) \\
\midrule
\multicolumn{2}{@{}l}{\emph{Selection and map (Secs.~\ref{sec:selection} and~\ref{sec:aggregation} of the main paper, Sec.~\ref{app:implementation})}} \\
$n_t$, $s_t^{(i)}$ & number of heads with $s_t^{\ell,h}>0$ and the $i$-th largest positive score \\
$\rho_t^{\ell,h}$ & rank of head $(\ell,h)$ among the positive scores ($1$ = highest) \\
$u_i$, $v_i$ & normalized rank and normalized score of the rank--score curve \\
$i^\star$, $S_t$ & knee index and selected head set \\
$K$ & fixed head budget of the Top-$K$ selection (Sec.~\ref{app:budget}) \\
$P_j^{\ell,h}$, $\widetilde P_j^{\ell,h}$ & positive attention gradient $\lfloor G_j^{\ell,h}\rfloor_+$, before and after the $3\times3$ median \\
$\mathcal N(j)$ & $3\times3$ neighborhood of $j$ on the visual-token grid (replicate padding) \\
$\boldsymbol\omega_t$ & rank weights of the heads in $S_t$, decreasing linearly with $\rho_t^{\ell,h}$ \\
$\omega_t^{\ell,h}$ & weight of head $(\ell,h)\in S_t$, $2(|S_t|+1-\rho_t^{\ell,h})/(|S_t|(|S_t|+1))$ \\
$\operatorname{median}_{\boldsymbol\omega_t}$ & weighted median across the heads in $S_t$ \\
$M_t$, $M_{t,j}$, $\widehat M_{t,j}$ & heatmap of $y_t$, its value at visual token $j$ and its value after min--max normalization \\
\midrule
\multicolumn{2}{@{}l}{\emph{Evaluation (Secs.~\ref{sec:funcwords} and~\ref{sec:experiments} of the main paper)}} \\
$I$, $I_b$ & input image, and the image with a fraction $b$ of its pixels replaced \\
$b$ & perturbed fraction, $b\in\{0,0.1,\dots,0.9\}$ \\
$\mathcal T$ & token positions over which a log-probability change is averaged \\
$\delta_{\mathcal T}(I_b)$ & mean log-probability change at the positions $\mathcal T$ from $I$ to $I_b$ \\
$\bar M_t$ & ground-truth segmentation mask of the object named by word $y_t$ \\
$\mathcal I_{\bar M_t}\subset\mathcal I$ & visual tokens covered by $\bar M_t$ \\
$p_\theta^{\setminus(S,R)}$ & output probability after blocking every head in $S$ from $R$ at the predicting row \\
$\delta_t(S_t,R)$ & drop in the final log-probability of $y_t$ when the heads in $S_t$ are blocked from $R$ \\
$y_n$, $\bar M_n$ & noun introduced by the function word $y_t$ ($n>t$) and the ground-truth mask of its object \\
$\mathcal F$ & positions of the other function words of the caption, each introducing a different object \\
$I_b^{t}$, $I_b^{t'}$ & $I_b$ with the highest-ranked cells of the map of $y_t$ or of $y_{t'}$ ($t'\in\mathcal F$) replaced \\
$\mathrm{FWP}(b)$, FWP-AUC & function-word perturbation at fraction $b$ and its normalized area over $b$ \\
$\mathcal O_t$ & masks of the other objects introduced by function words in the caption \\
$\mathrm{FWS}_t$ & function-word segmentation score \\
\bottomrule
\end{tabular}
\caption{Symbols used in the paper and the supplementary material.}
\label{tab:app-notation}
\end{table*}

\section{Derivation of the Head-Ranking Score}
\label{app:knockout}

We first derive how an attention knockout~\cite{geva-dissecting-recall-2023,kaduri-whats-in-the-image-2025} changes a head's attention weights. We then use that exact change
to obtain a first-order approximation of the knockout's effect on the target word's layer logit
(Eq.~\eqref{eq:app-knockout-approximation}) and use the approximation to define the head-ranking score
(Eq.~\eqref{eq:app-normalized-score}). Throughout this section the step $t$, layer $\ell$ and head $h$ are fixed.

\subsection{Attention Knockout and Renormalization}
\label{app:renorm}

Blocking a head from a set of visual tokens sets their attention weights to zero. The softmax then redistributes the
removed weight in proportion to the original weights of the remaining keys. To derive this renormalization, we start
with the attention computation before any tokens are blocked. Each head computes pre-softmax attention scores from its query and key projections. Its attention weights are the
softmax of these scores over the keys:
\begin{equation}
\mathbf e^{\ell,h}=\frac{\mathbf Q^{\ell,h}{\mathbf K^{\ell,h}}^{\top}}{\sqrt{d_k}},\qquad
\mathbf A^{\ell,h}=\operatorname{softmax}\bigl(\mathbf e^{\ell,h}\bigr),
\label{eq:app-attention-matrix}
\end{equation}
where $\mathbf e^{\ell,h}$ and $\mathbf A^{\ell,h}$ are the scores and the weights of head $(\ell,h)$,
$\mathbf Q^{\ell,h}$ and $\mathbf K^{\ell,h}$ are its query and key projections, and $d_k$ is the key dimension per head.
For the knockout, we only need the row used to predict the target word. At the predicting row, the attention weight to each key is its exponentiated score divided by the sum over all keys:
\begin{equation}
A_k^{\ell,h}=\frac{\exp(e_k^{\ell,h})}{\sum_{k'\in\mathcal K}\exp(e_{k'}^{\ell,h})},
\label{eq:app-attention}
\end{equation}
where $e_k^{\ell,h}$ and $A_k^{\ell,h}$ are the score and the weight to key $k\in\mathcal K$. Rearranging this expression lets us write each numerator in terms of the original attention weight and the original
denominator, as $\exp(e_k^{\ell,h})=A_k^{\ell,h}\sum_{k'\in\mathcal K}\exp(e_{k'}^{\ell,h})$.

Knocking out a visual region $R\subset\mathcal I$ sets $e_j^{\ell,h}=-\infty$ for $j\in R$ and leaves every other
score unchanged. The blocked numerators become $\exp(-\infty)=0$, so the denominator loses exactly the terms for the blocked keys. With $\exp(e_j^{\ell,h})=A_j^{\ell,h}\sum_{k'\in\mathcal K}\exp(e_{k'}^{\ell,h})$, the
softmax denominator after the knockout is
\begin{align}
&\sum_{k\in\mathcal K\setminus R}\exp(e_k^{\ell,h})\nonumber\\
&\quad=\sum_{k'\in\mathcal K}\exp(e_{k'}^{\ell,h})-\sum_{j\in R}\exp(e_j^{\ell,h})\nonumber\\
&\quad=\sum_{k'\in\mathcal K}\exp(e_{k'}^{\ell,h})-\sum_{j\in R}A_j^{\ell,h}\sum_{k'\in\mathcal K}\exp(e_{k'}^{\ell,h})\nonumber\\
&\quad=\Bigl(1-\sum_{j\in R}A_j^{\ell,h}\Bigr)\sum_{k'\in\mathcal K}\exp(e_{k'}^{\ell,h}).
\label{eq:app-softmax-denom}
\end{align}
A remaining key $k\in\mathcal K\setminus R$ keeps its numerator $\exp(e_k^{\ell,h})$ and gets this new denominator.
Since $\exp(e_k^{\ell,h})/\sum_{k'\in\mathcal K}\exp(e_{k'}^{\ell,h})=A_k^{\ell,h}$, the attention weights after the knockout are
\begin{equation}
A_k^{\ell,h,\setminus R}=
\begin{cases}
0, & k\in R,\\[2pt]
\dfrac{A_k^{\ell,h}}{1-\sum_{j\in R}A_j^{\ell,h}}, & k\in\mathcal K\setminus R,
\end{cases}
\label{eq:app-renorm}
\end{equation}
where $A_k^{\ell,h,\setminus R}$ is the attention weight to key $k$ after head $(\ell,h)$ is blocked from $R$.
The new denominator scales up the weights of all remaining keys by the same factor, including the weights of text
tokens. Since the blocked keys are visual and at least one text key remains, $\sum_{j\in R}A_j^{\ell,h}<1$ for finite scores. 

To express the knockout's effect on the layer logit, we will use the difference between the original and knocked-out weights.
A blocked key $j\in R$ loses its whole weight, so its change is
$A_j^{\ell,h}-0=A_j^{\ell,h}$. For a remaining key $k\in\mathcal K\setminus R$, the change
in attention after blocking $R$ is
\begin{align}
A_k^{\ell,h}-A_k^{\ell,h,\setminus R}
&=A_k^{\ell,h}-\frac{A_k^{\ell,h}}{1-\sum_{j\in R}A_j^{\ell,h}}\nonumber\\
&=A_k^{\ell,h}\,\frac{\bigl(1-\sum_{j\in R}A_j^{\ell,h}\bigr)-1}{1-\sum_{j\in R}A_j^{\ell,h}}\nonumber\\
&=-\frac{\sum_{j\in R}A_j^{\ell,h}}{1-\sum_{j\in R}A_j^{\ell,h}}\,A_k^{\ell,h}.
\label{eq:app-change}
\end{align}
The blocked keys lose a total weight of $\sum_{j\in R}A_j^{\ell,h}$. Before the knockout, the remaining keys have
total weight $1-\sum_{j\in R}A_j^{\ell,h}$, so Eq.~\eqref{eq:app-change} shows that they gain exactly
$\sum_{j\in R}A_j^{\ell,h}$. The losses and gains cancel, so the knocked-out row still sums to one.

\subsection{First-Order Approximation}
\label{app:taylor}

The renormalization gives the exact change to the attention row. To estimate the effect of that change on the layer
logit, we use the gradient at the original row. We read the layer logit of the target word with the logit lens~\cite{nostalgebraist-logit-lens-2020}. It projects the representation of the
predicting position after layer $\ell$ onto the vocabulary:
\begin{equation}
\hat o^{\ell}(y_t)=\bigl[\mathbf W_{\mathrm U}\operatorname{Norm}(\mathbf z_{-1}^{\ell})\bigr]_{y_t},
\label{eq:app-logit-lens}
\end{equation}
where $\hat o^{\ell}(y_t)$ is the layer logit of $y_t$, $\mathbf z_{-1}^{\ell}$ is the representation of the predicting
position after layer $\ell$, $\operatorname{Norm}$ is the model's final normalization and $\mathbf W_{\mathrm U}$ is the
unembedding matrix.
To isolate the effect of the knockout, we treat $\hat o^{\ell}(y_t)$ as a function of the attention row of head
$(\ell,h)$ and keep all other inputs, parameters and heads fixed. The attention knockout score is the layer
logit at the original row minus the layer logit at the knocked-out row:
\begin{equation}
\Delta^{\ell,h}(R)=\hat o^{\ell}(y_t)-\hat o^{\ell,\setminus(h,R)}(y_t),
\label{eq:app-knockout-score}
\end{equation}
where $\hat o^{\ell,\setminus(h,R)}(y_t)$ is the layer logit after head $(\ell,h)$ is blocked from $R$.
A first-order Taylor expansion around the original row approximates the layer logit after the knockout as
\begin{equation}
\hat o^{\ell,\setminus(h,R)}(y_t)\approx\hat o^{\ell}(y_t)+\sum_{k\in\mathcal K}G_k^{\ell,h}
\bigl(A_k^{\ell,h,\setminus R}-A_k^{\ell,h}\bigr),
\label{eq:app-taylor}
\end{equation}
where $G_k^{\ell,h}=\partial\hat o^{\ell}(y_t)/\partial A_k^{\ell,h}$ is the attention-gradient for key $k$, taken at the
original row. Subtracting this from
the original layer logit in Eq.~\eqref{eq:app-knockout-score} gives the first-order approximation of the knockout score:
\begin{equation}
\Delta^{\ell,h}(R)\approx\sum_{k\in\mathcal K}G_k^{\ell,h}\bigl(A_k^{\ell,h}-A_k^{\ell,h,\setminus R}\bigr).
\label{eq:app-first-order-drop}
\end{equation}

The weights change differently for the two groups of keys. Each blocked key $j\in R$ loses its weight $A_j^{\ell,h}$,
while each remaining key $k\in\mathcal K\setminus R$ gains weight according to Eq.~\eqref{eq:app-change}. The remaining
weights all change in the same proportion, so their common factor can be taken outside the sum over $k$.
This splits the approximation into two terms:
\begin{align}
\Delta^{\ell,h}(R)
&\approx\underbrace{\sum_{j\in R}A_j^{\ell,h}G_j^{\ell,h}}_{\text{removed attention}}\nonumber\\
&\quad-\underbrace{\frac{\sum_{j\in R}A_j^{\ell,h}}{1-\sum_{j\in R}A_j^{\ell,h}}\sum_{k\in\mathcal K\setminus R}A_k^{\ell,h}G_k^{\ell,h}}_{\text{redistributed attention}}.
\label{eq:app-separated-knockout}
\end{align}
The first term measures the effect of removing attention from $R$. The second accounts for the effect of giving that
same attention mass to the remaining keys.

To combine the two terms, we split the attention-weighted mean gradient over all keys into the same two groups of
keys:
\begin{equation}
\bar G^{\ell,h}=\sum_{j\in R}A_j^{\ell,h}G_j^{\ell,h}+\sum_{k\in\mathcal K\setminus R}A_k^{\ell,h}G_k^{\ell,h},
\label{eq:app-gbar-split}
\end{equation}
where $\bar G^{\ell,h}$ is the attention-weighted mean gradient of the original row. This lets us express the remaining
keys' contribution through the mean over the whole row and the blocked keys' contribution, as
$\sum_{k\in\mathcal K\setminus R}A_k^{\ell,h}G_k^{\ell,h}=\bar G^{\ell,h}-\sum_{j\in R}A_j^{\ell,h}G_j^{\ell,h}$.
Inserting this into Eq.~\eqref{eq:app-separated-knockout} and collecting the two terms in $\sum_{j\in R}A_j^{\ell,h}G_j^{\ell,h}$ gives the approximation of the knockout score in closed form:
\begin{align}
\Delta^{\ell,h}(R)
&\approx\sum_{j\in R}A_j^{\ell,h}G_j^{\ell,h}\nonumber\\
&\quad-\frac{\sum_{j\in R}A_j^{\ell,h}}{1-\sum_{j\in R}A_j^{\ell,h}}
\Bigl(\bar G^{\ell,h}-\sum_{j\in R}A_j^{\ell,h}G_j^{\ell,h}\Bigr)\nonumber\\
&=\Bigl(1+\frac{\sum_{j\in R}A_j^{\ell,h}}{1-\sum_{j\in R}A_j^{\ell,h}}\Bigr)\sum_{j\in R}A_j^{\ell,h}G_j^{\ell,h}\nonumber\\
&\quad-\frac{\sum_{j\in R}A_j^{\ell,h}}{1-\sum_{j\in R}A_j^{\ell,h}}\,\bar G^{\ell,h}\nonumber\\
&=\frac{\sum_{j\in R}A_j^{\ell,h}G_j^{\ell,h}-\bar G^{\ell,h}\sum_{j\in R}A_j^{\ell,h}}{1-\sum_{j\in R}A_j^{\ell,h}}
\nonumber\\
&=\frac{\sum_{j\in R}A_j^{\ell,h}\bigl(G_j^{\ell,h}-\bar G^{\ell,h}\bigr)}{1-\sum_{j\in R}A_j^{\ell,h}}.
\label{eq:app-knockout-approximation}
\end{align}
The step to the third equality uses $1+\frac{x}{1-x}=\frac{1}{1-x}$ with $x=\sum_{j\in R}A_j^{\ell,h}$ and puts both terms over the
common denominator. The last equality moves the constant $\bar G^{\ell,h}$ inside the sum over $R$. The numerator
then compares each blocked key's gradient with the original row's mean, weighted by the key's original attention.
The denominator is the original attention mass on the remaining keys.

Since the renormalization and the algebra are exact, the accuracy of the result depends on the first-order expansion in
Eq.~\eqref{eq:app-taylor}. The expansion uses the gradient at the original row to estimate the effect of a finite
change in attention. Sec.~\ref{app:validity} measures the accuracy
of this approximation.

We can also express the approximation directly as the effect of reweighting the original gradients. After the
knockout, the attention-weighted mean gradient over the remaining keys follows from Eqs.~\eqref{eq:app-renorm}
and~\eqref{eq:app-gbar-split}:
\begin{equation}
\bar G^{\ell,h}_{\setminus R}=\sum_{k\in\mathcal K\setminus R}A_k^{\ell,h,\setminus R}G_k^{\ell,h}
=\frac{\bar G^{\ell,h}-\sum_{j\in R}A_j^{\ell,h}G_j^{\ell,h}}{1-\sum_{j\in R}A_j^{\ell,h}},
\label{eq:app-gbar-after}
\end{equation}
where $\bar G^{\ell,h}_{\setminus R}$ is the mean of the original gradients under the knocked-out attention weights.
The difference between the old and the new mean gradient, over the common denominator, is
\begin{equation}
\bar G^{\ell,h}-\bar G^{\ell,h}_{\setminus R}
=\frac{\sum_{j\in R}A_j^{\ell,h}G_j^{\ell,h}-\bar G^{\ell,h}\sum_{j\in R}A_j^{\ell,h}}{1-\sum_{j\in R}A_j^{\ell,h}},
\label{eq:app-mean-difference}
\end{equation}
which is the third equality of Eq.~\eqref{eq:app-knockout-approximation}. The predicted logit drop is the original
mean gradient minus the mean obtained by reweighting those same gradients.

\subsection{The Token's Score as a Softmax Gradient}
\label{app:suppression}

The numerator of the knockout approximation contains the term $A_j^{\ell,h}(G_j^{\ell,h}-\bar G^{\ell,h})$ for each
blocked token. For a visual token, that same term describes the local change in the layer logit as the token's
pre-softmax attention score increases. We now compute the derivative of the layer logit with
respect to the pre-softmax score of visual token $j$. The score affects the layer logit only through the attention
weights, so by the chain rule
\begin{equation}
\frac{\partial\hat o^{\ell}(y_t)}{\partial e_j^{\ell,h}}
=\sum_{k\in\mathcal K}\frac{\partial\hat o^{\ell}(y_t)}{\partial A_k^{\ell,h}}\,
\frac{\partial A_k^{\ell,h}}{\partial e_j^{\ell,h}}
=\sum_{k\in\mathcal K}G_k^{\ell,h}\,\frac{\partial A_k^{\ell,h}}{\partial e_j^{\ell,h}}.
\label{eq:app-chain}
\end{equation}
The softmax derivative accounts for how changing one score affects the entire row. From Eq.~\eqref{eq:app-attention},
the shared denominator depends on $e_j^{\ell,h}$
through its term $\exp(e_j^{\ell,h})$, so its derivative with respect to $e_j^{\ell,h}$ is $\exp(e_j^{\ell,h})$. For $k=j$, the quotient rule gives the derivative of
the weight of token $j$ with respect to its own score:
\begin{align}
\frac{\partial A_j^{\ell,h}}{\partial e_j^{\ell,h}}
&=\frac{\exp(e_j^{\ell,h})\sum_{k'\in\mathcal K}\exp(e_{k'}^{\ell,h})-\exp(e_j^{\ell,h})^2}{\bigl(\sum_{k'\in\mathcal K}\exp(e_{k'}^{\ell,h})\bigr)^2}\nonumber\\
&=A_j^{\ell,h}-\bigl(A_j^{\ell,h}\bigr)^2.
\end{align}
For $k\neq j$, only the denominator depends on $e_j^{\ell,h}$, and the derivative of the
weight of key $k$ is
\begin{equation}
\frac{\partial A_k^{\ell,h}}{\partial e_j^{\ell,h}}
=-\frac{\exp(e_k^{\ell,h})\exp(e_j^{\ell,h})}{\bigl(\sum_{k'\in\mathcal K}\exp(e_{k'}^{\ell,h})\bigr)^2}=-A_k^{\ell,h}A_j^{\ell,h}.
\end{equation}
Both cases are $\partial A_k^{\ell,h}/\partial e_j^{\ell,h}=A_k^{\ell,h}(\mathbf 1[k=j]-A_j^{\ell,h})$. Inserting this into
Eq.~\eqref{eq:app-chain} gives the derivative in terms of the attention-gradients:
\begin{equation}
\begin{split}
\frac{\partial\hat o^{\ell}(y_t)}{\partial e_j^{\ell,h}}
&=\sum_{k\in\mathcal K}G_k^{\ell,h}A_k^{\ell,h}\bigl(\mathbf 1[k=j]-A_j^{\ell,h}\bigr)\\
&=A_j^{\ell,h}G_j^{\ell,h}-A_j^{\ell,h}\sum_{k\in\mathcal K}A_k^{\ell,h}G_k^{\ell,h}\\
&=A_j^{\ell,h}\bigl(G_j^{\ell,h}-\bar G^{\ell,h}\bigr)=C_j^{\ell,h},
\end{split}
\label{eq:app-contribution-identity}
\end{equation}
where $C_j^{\ell,h}$ is the token's score of visual token $j$.
The indicator keeps only the term $k=j$ of the first sum, and the second sum is $\bar G^{\ell,h}$. The centering by
$\bar G^{\ell,h}$ captures the effect of the shared softmax denominator, which shifts attention across the row as
one score rises. $C_j^{\ell,h}>0$ means that raising the attention score of visual token $j$, with all other scores
fixed, initially raises the layer logit of $y_t$.

\subsection{Tokens with Positive Scores}
\label{app:positive-region}

The blocked region $R$ determines what the score of a head measures. Blocking the whole image ($R=\mathcal I$)
measures how much the head's use of the entire image contributes to the target word $y_t$. Its numerator
$\sum_{j\in\mathcal I}C_j^{\ell,h}$ then sums positive and negative token scores, which can cancel each other.
A head with large positive scores on the object and large negative scores on the background can therefore receive
a score close to zero. We thus block the set $R_+^{\ell,h}$ of visual tokens with positive scores, so that the
score reflects only the visual evidence that supports $y_t$. Sec.~\ref{app:alt-regions} compares both choices.

Blocking $R_+^{\ell,h}$ also maximizes the numerator of Eq.~\eqref{eq:app-knockout-approximation} over all
$R\subseteq\mathcal I$. By the definition of the token's score, this numerator equals
$\sum_{j\in R}A_j^{\ell,h}(G_j^{\ell,h}-\bar G^{\ell,h})=\sum_{j\in R}C_j^{\ell,h}$, and splitting $R$ by the sign
of $C_j^{\ell,h}$ gives
\begin{align}
&\sum_{j\in R}C_j^{\ell,h}\nonumber\\
&\quad=\sum_{j\in R,\,C_j^{\ell,h}>0}C_j^{\ell,h}+\sum_{j\in R,\,C_j^{\ell,h}\le0}C_j^{\ell,h}\nonumber\\
&\quad\le\sum_{j\in R,\,C_j^{\ell,h}>0}C_j^{\ell,h}\nonumber\\
&\quad\le\sum_{j\in\mathcal I,\,C_j^{\ell,h}>0}C_j^{\ell,h}=\sum_{j\in\mathcal I}\lfloor C_j^{\ell,h}\rfloor_+,
\label{eq:app-positive-bound}
\end{align}
where $\lfloor x\rfloor_+=\max(x,0)$.
The first inequality drops terms that are zero or negative. The second adds the positive terms for visual tokens
outside $R$. Both inequalities become equalities for $R=R_+^{\ell,h}$, which consists of the visual tokens with
positive scores. 

\subsection{The Head-Ranking Score}
\label{app:normalized-score}

Having selected the visual tokens with positive scores, we evaluate the knockout approximation on that set to rank
the head. The resulting score estimates the drop in the target word's layer logit after blocking those tokens and
redistributing their attention. On $R=R_+^{\ell,h}$, the numerator of
Eq.~\eqref{eq:app-knockout-approximation} is $\sum_{j\in\mathcal I}\lfloor C_j^{\ell,h}\rfloor_+$ by
Eq.~\eqref{eq:app-positive-bound}, and the denominator is $1-\sum_{j\in R_+^{\ell,h}}A_j^{\ell,h}$. This gives the head-ranking score:
\begin{equation}
s_t^{\ell,h}=
\frac{\sum_{j\in\mathcal I}\lfloor C_j^{\ell,h}\rfloor_+}{1-\sum_{j\in R_+^{\ell,h}}A_j^{\ell,h}}
\approx\Delta^{\ell,h}(R_+^{\ell,h}),
\label{eq:app-normalized-score}
\end{equation}
where $s_t^{\ell,h}$ is the head-ranking score of head $(\ell,h)$ for the target word $y_t$.
We select $R_+^{\ell,h}$ by maximizing the numerator. Evaluating the score on that set then uses the denominator to
account for the renormalization of the remaining attention weights.

\section{Head Selection and Map Construction}
\label{app:implementation}
\wish computes the score $s_t^{\ell,h}$ of all $L \times H$ heads and ranks the heads with $s_t^{\ell,h}>0$ jointly across
layers. 

\subsection{Adaptive Head Selection}
\label{app:knee}

The selected set $S_t$ contains all heads
up to the knee of the sorted scores, where the steep drop turns into the flat tail. We clamp the
denominator $1-\sum_{j\in R_+^{\ell,h}}A_j^{\ell,h}$ of the score at $10^{-6}$ to avoid division by zero. We normalize the score and the rank to form the score-rank curve with $u_i=\frac{i-1}{n_t-1}$ and
$v_i=\frac{s_t^{(i)}-s_t^{(n_t)}}{s_t^{(1)}-s_t^{(n_t)}}$. The curve $(u_i,v_i)$ joins $(0,1)$ and $(1,0)$. The knee
$i^\star=\argmax_{1\le i\le n_t}(1-u_i-v_i)$ is the point farthest below the chord, measured vertically. 
The selected head set keeps the top $i^\star$ heads (Eq.~\eqref{eq:wish-selected-heads}). 
If several indices attain the maximum, the
smallest is used.

Every explained word in our evaluations has at least $897$ heads with $s_t^{\ell,h}>0$, out of $1{,}024$--$1{,}280$.
We keep at least three heads per token. This minimum affected one of
roughly $100{,}000$ explained tokens. Tab.~\ref{tab:app-knee-size}
reports the selected set size $|S_t|$ by model and dataset.

\begin{table}[t]
\centering
\small
\setlength{\tabcolsep}{5pt}
\begin{tabular}{@{}clrrrr@{}}
\toprule
 & Model & Mean & Std & Min & Max \\
\midrule
\multirow{6}{*}{\rotatebox[origin=c]{90}{PASCAL VOC}} & LLaVA-1.5-7B & 58.8 & 17.0 & 16 & 136 \\
 & BakLLaVA & 70.7 & 19.5 & 24 & 140 \\
 & Qwen3-VL-8B & 68.6 & 21.5 & 24 & 202 \\
 & GLM-4.6V-Flash & 68.7 & 23.5 & 22 & 185 \\
 & LLaVA-OV-1.5-8B & 61.8 & 18.6 & 27 & 181 \\
 & LLaVA-OV-2-8B & 62.5 & 19.8 & 23 & 165 \\
\midrule
\multirow{6}{*}{\rotatebox[origin=c]{90}{COCO}} & LLaVA-1.5-7B & 56.6 & 22.0 & 13 & 173 \\
 & BakLLaVA & 68.7 & 21.8 & 20 & 147 \\
 & Qwen3-VL-8B & 68.3 & 23.3 & 13 & 205 \\
 & GLM-4.6V-Flash & 68.0 & 27.4 & 13 & 172 \\
 & LLaVA-OV-1.5-8B & 64.6 & 22.8 & 14 & 165 \\
 & LLaVA-OV-2-8B & 66.5 & 24.5 & 8 & 198 \\
\midrule
\multirow{6}{*}{\rotatebox[origin=c]{90}{ImageNet}} & LLaVA-1.5-7B & 73.9 & 25.1 & 11 & 192 \\
 & BakLLaVA & 85.5 & 26.8 & 9 & 198 \\
 & Qwen3-VL-8B & 83.9 & 37.9 & 8 & 235 \\
 & GLM-4.6V-Flash & 100.0 & 36.8 & 9 & 251 \\
 & LLaVA-OV-1.5-8B & 95.2 & 35.8 & 12 & 240 \\
 & LLaVA-OV-2-8B & 93.1 & 31.7 & 10 & 231 \\
\midrule
\multirow{6}{*}{\rotatebox[origin=c]{90}{VQAv2}} & LLaVA-1.5-7B & 69.1 & 22.8 & 13 & 153 \\
 & BakLLaVA & 81.5 & 25.7 & 14 & 178 \\
 & Qwen3-VL-8B & 87.2 & 36.4 & 4 & 223 \\
 & GLM-4.6V-Flash & 85.6 & 37.4 & 14 & 214 \\
 & LLaVA-OV-1.5-8B & 94.0 & 36.3 & 3 & 200 \\
 & LLaVA-OV-2-8B & 86.6 & 36.2 & 6 & 217 \\
\bottomrule
\end{tabular}
\caption{Size of the adaptively selected head set $|S_t|$ per explained token. Reported statistics are mean, standard deviation (Std),
minimum (Min) and maximum (Max). Every row is the statistics of the size of $|S_t|$ of one model on one dataset.}
\label{tab:app-knee-size}
\end{table}

\subsection{Fixed Top-$K$ Head Selection}
\label{app:budget}

We also evaluate selecting $S_t$ as the $K$ highest-scoring positive heads. We choose $K$ on the captions of the PASCAL VOC 2012 \emph{training} images. These are 869--1{,}104 images with an object
word per model, disjoint from all evaluation images. We sweep $K\in\{8,16,32,64,128,256\}$ and all positive heads, and score Obj-IoU, positive deletion and insertion (Fig.~\ref{fig:app-budget}). This analysis shows that the best tuned budget varies per model. Comparing the six-model mean budgets gives $K=64$ for the highest Obj-IoU ($36.74$) and the highest positive deletion ($2.260$), and insertion prefers $K=128$ slightly ($0.500$). Thus, we use $K=64$ as a tuned fixed budget. 
The knee set matches the best fixed budget on all three metrics ($36.5$, $2.259$ and $0.506$) without choosing $K$.

\subsection{Spatial Filtering and Head Aggregation}
\label{app:aggregation}

Each selected head's map is denoised with a $3\times3$ median. The maps are then combined by a weighted median in
which higher-ranked heads count more. Selection is fixed before filtering. The filter and the median
therefore change only the map and not which heads are used. For each selected head, the positive attention gradient at each visual token is replaced by its $3\times3$ median as in Eq.~\eqref{eq:wish-head-map-filtered} of the main paper.
With $|S_t|$ selected heads, the weights $\boldsymbol\omega_t$ of the weighted median in Eq.~\eqref{eq:wish-head-median} of the main paper are
\begin{equation}
\omega_t^{\ell,h}=\frac{2(|S_t|+1-\rho_t^{\ell,h})}{|S_t|(|S_t|+1)},\qquad (\ell,h)\in S_t.
\label{eq:app-head-weights}
\end{equation}
Finally, the heatmap $M_t$ is min--max normalized over the visual tokens:
\begin{equation}
\widehat M_{t,j}=\frac{M_{t,j}-\min_{j'\in\mathcal I}M_{t,j'}}{\max_{j'\in\mathcal I}M_{t,j'}-\min_{j'\in\mathcal I}M_{t,j'}},
\label{eq:app-map-normalization}
\end{equation}
where $\widehat M_{t,j}$ is the normalized value. Every map is then upsampled bilinearly to the image size for evaluation. A constant map ($\max=\min$) is set to zero.

\section{Validity of the Attention Knockout \texorpdfstring{\mbox{Approximation}}{Approximation}}
\label{app:validity}

The head-ranking score (Eq.~\eqref{eq:app-normalized-score}) estimates how much the layer logit
$\hat o^{\ell}(y_t)$ drops when one head is blocked from $R_+^{\ell,h}$. We compare it with exact knockouts
that block the same head and read the logit at the same layer. This differs from the knockout of Sec.~\ref{sec:head_selection} of
the main paper, which blocks a set of heads jointly and reads the final log-probability of $y_t$. 
\subsection{Accuracy Against Exact Knockouts}
\label{app:exact-agreement}

To test how well the score ranks heads, we compare it with the quantity it approximates, namely, the logit drop measured after an exact knockout of each head. For each PASCAL VOC object word, we block each head's attention to $R_+^{\ell,h}$ in turn (Eq.~\eqref{eq:app-renorm}), recompute layer $\ell$, and measure the drop in $\hat o^{\ell}(y_t)$.

\begin{figure}[t]
\centering
\includegraphics[width=\linewidth]{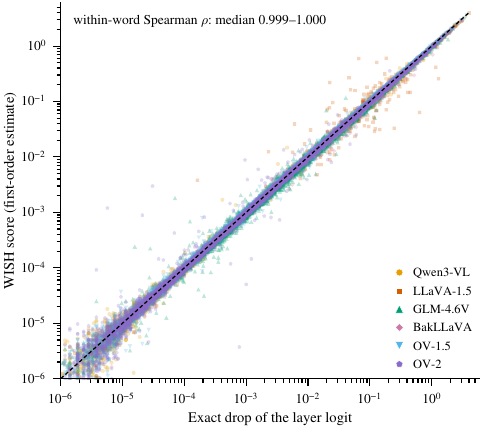}
\caption{\wish head-ranking score (Eq.~\eqref{eq:app-normalized-score}) versus the exact drop in the layer logit,
for 20{,}000 random heads with a positive score and drop per model (PASCAL VOC object words). Dashed: $y=x$. Top
left: median within-word Spearman correlation over all heads, as in Tab.~\ref{tab:app-exact}.}
\label{fig:eq4-exact}
\end{figure}

The score and the measured drops produce nearly identical head rankings when computed separately for each word.
Fig.~\ref{fig:eq4-exact} plots the score of each head against its exact drop, and Tab.~\ref{tab:app-exact} gives,
per model, the rank and linear correlation between the two within each word and the overlap of the head sets they
select.
Across models, the median within-word Spearman $\rho$ ranges from $0.9994$ to $0.9999$, with 924--1{,}137 words per model.

\begin{figure*}[t]
\centering
\includegraphics[width=\textwidth]{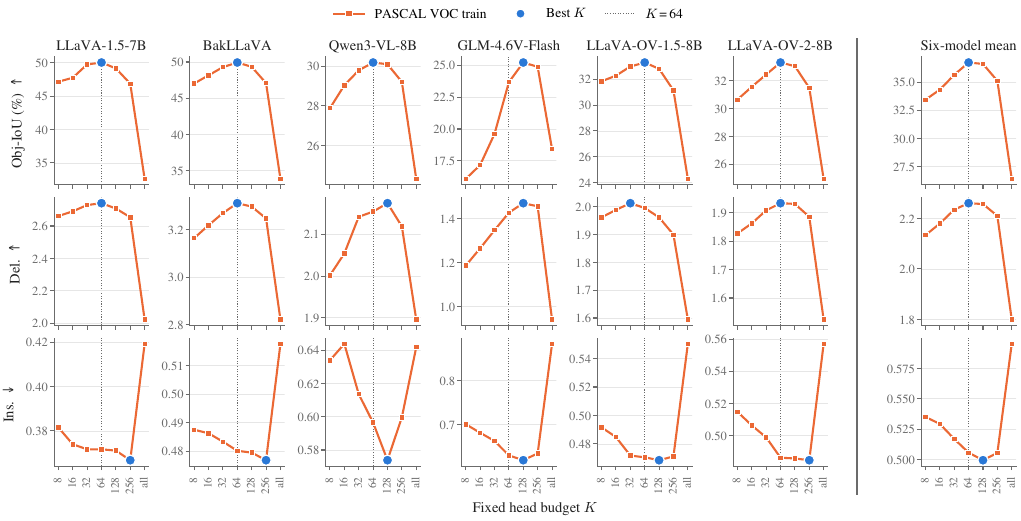}
\caption{\wish with a fixed head budget $K$ (Sec.~\ref{app:budget}). Obj-IoU, positive deletion (Del.) and insertion
(Ins.) on the PASCAL VOC 2012 training set. Dotted: $K=64$. Blue dot is best $K$ per model. ``all'' denotes every head with $s_t^{\ell,h}>0$.}
\label{fig:app-budget}
\end{figure*}

\begin{table}[t]
\centering
\small
\setlength{\tabcolsep}{2.5pt}
\begin{tabular}{@{}lrccc@{}}
\toprule
Model & Words & Spearman & Pearson & Knee $J$ \\
\midrule
LLaVA-1.5 & 1033 & 0.9996 (0.9883) & 0.988 (0.967) & 0.89 \\
BakLLaVA & 1008 & 0.9999 (0.9995) & 0.999 (0.997) & 0.92 \\
Qwen3-VL & 924 & 0.9998 (0.9998) & 0.999 (0.999) & 0.92 \\
GLM-4.6V & 1082 & 0.9994 (0.9991) & 0.998 (0.997) & 0.88 \\
OV-1.5 & 1137 & 0.9998 (0.9998) & 0.999 (0.999) & 0.92 \\
OV-2 & 1106 & 0.9997 (0.9996) & 0.999 (0.999) & 0.91 \\
\bottomrule
\end{tabular}
\caption{Agreement between the \wish head-ranking score and the exact drop in the layer logit on PASCAL VOC object
words. Spearman and Pearson: median over words of the correlation across all heads, with the 10th percentile in
parentheses. Knee $J$: mean Jaccard index between the knee sets selected by the score and by the exact drops.}
\label{tab:app-exact}
\end{table}
The close agreement in head rankings carries over to downstream tasks like localization and perturbation.
Using the exact drops to rank heads, PASCAL VOC localization metrics such as Obj-IoU change by $-0.06$ (Tab.~\ref{tab:app-knockout-rank}) and perturbation log AUC by at most $0.012$ in absolute value (Tab.~\ref{tab:app-exact-perturb}).

\begin{table}[t]
\centering
\small
\begin{tabular}{@{}lcccc@{}}
\toprule
& \multicolumn{2}{c}{Knee} & \multicolumn{2}{c}{Top-64} \\
\cmidrule(lr){2-3}\cmidrule(l){4-5}
Model & Del. & Ins. & Del. & Ins. \\
\midrule
LLaVA-1.5 & $+.009$ & $-.002$ & $+.012$ & $+.000$ \\
BakLLaVA & $+.003$ & $-.000$ & $+.002$ & $+.000$ \\
Qwen3-VL & $-.000$ & $+.007$ & $-.003$ & $-.005$ \\
GLM-4.6V & $-.003$ & $+.003$ & $+.000$ & $+.000$ \\
OV-1.5 & $+.004$ & $+.001$ & $+.001$ & $+.001$ \\
OV-2 & $+.009$ & $-.001$ & $+.004$ & $+.001$ \\
\bottomrule
\end{tabular}
\caption{Perturbation log AUC with heads ranked by the exact drop in the layer logit, minus that of \wish, with the
same selection rule and map construction. PASCAL VOC object words. Del.\ and Ins.\ as in Tab.~\ref{tab:perturbation} of the main paper.}
\label{tab:app-exact-perturb}
\end{table}

\subsection{Blocked Region and Readout}
\label{app:knockout-rankings}
\label{app:alt-regions}

Tab.~\ref{tab:app-knockout-rank} tests which readout and which blocked region give the best maps. Instead of
\wish's approximated drop, we rank heads by the exact drop measured after blocking one head at a time, and keep
\wish's knee rule and map construction. The drop is read either in the layer logit of the head's own layer or in the
final log-probability of the model's output. The blocked region is $R_+^{\ell,h}$, the whole image $\mathcal I$ or
the annotated object $\mathcal I_{\bar M_t}$. Each row reports the change in Obj-IoU relative to \wish. With the layer logit, blocking $R_+^{\ell,h}$ matches \wish's maps ($-0.06$), and blocking the whole image lowers
Obj-IoU by $1.67$. Blocking the annotated object raises it by $8.27$, but this region requires the segmentation
ground-truth mask. With the final log-probability, Obj-IoU falls by $7.8$ to $16.1$ for every region. The final
log-probability does not necessarily show where the head reads its evidence.
\begin{table}[t]
\centering
\small
\setlength{\tabcolsep}{3.5pt}
\begin{tabular}{@{}lllr@{}}
\toprule
Drop & Readout & Blocked region & $\Delta$Obj-IoU \\
\midrule
Approximated & layer logit & $R_+^{\ell,h}$ & $0.00$ \\
Exact & layer logit & $R_+^{\ell,h}$ & $-0.06$ \\
Exact & layer logit & $\mathcal{I}$ & $-1.67$ \\
Exact & layer logit & $\mathcal I_{\bar M_t}$ (oracle) & $+8.27$ \\
Exact & final $\log p$ & $R_+^{\ell,h}$ & $-7.78$ \\
Exact & final $\log p$ & $\mathcal{I}$ & $-15.87$ \\
Exact & final $\log p$ & $\mathcal I_{\bar M_t}$ (oracle) & $-16.10$ \\
\bottomrule
\end{tabular}
\caption{Obj-IoU differences from \wish (first row), six-model means. Words: the PASCAL VOC object words of
Sec.~\ref{app:exact-agreement} for which every ranking selects at least three heads (845--1{,}106 per model). Oracle
region $\mathcal I_{\bar M_t}$: the annotated object mask.}
\label{tab:app-knockout-rank}
\end{table}

\section{Computational Cost}
\label{app:cost}
\label{app:cost-knockout}

We compare the head-ranking score with the exact knockouts it
approximates on one NVIDIA A100 PCIe 40\,GB GPU using each model's default precision.
Both methods start with a shared forward pass over the caption
prefix to cache each layer's input and the attention weights and
values used by the predicting row. The approximation uses one
backward pass per layer, from the layer logit to the head outputs.
For each exact knockout, we use the token scores to identify
$R_+^{\ell,h}$, block the head's attention to that region, and
recompute the predicting row through layer $\ell$
(Sec.~\ref{app:exact-agreement}).

We benchmark the object words from PASCAL VOC (Sec.~\ref{app:exact-agreement}).
For the single-word timings, we average over the total object words per model. For the batched timings, we use a batch of 1{,}024 words per model and average, and we report the time per row. If the number of object words per model is less than the batch (1{,}024 words) we repeat the first three words.

Tab.~\ref{tab:app-cost} shows that the time gap between the exact knockouts and the approximation grows with
batching. Exact knockouts take 1.4--1.5 times as long as the approximation for a single word and 10.7--12.7 times
as long in a batch of 1{,}024 words. Since a batch reads the model weights once for all its words, the recomputation
of one knockout per head accounts for a larger share of the runtime. Exact knockouts also require more GPU memory in
both settings, with 129--131\,MB instead of 5--12\,MB for a single word and 2.9--4.3\,GB instead of 0.54--0.64\,GB
in a batch.

\begin{table*}[t]
\centering
\small
\setlength{\tabcolsep}{3pt}
\begin{tabular}{@{}lccc@{}}
\toprule
Method and setting & Time per word (ms) & Time ratio & Peak extra GPU memory \\
\midrule
Approximation, single word inference & 65--107 & 1 & 5--12\,MB \\
Approximation, batched inference & 0.18--0.25 & 1 & 0.54--0.64\,GB \\
\midrule
Exact knockout, single word inference & 95--154 & 1.4--1.5 & 129--131\,MB \\
Exact knockout, batched inference & 2.0--3.2 & 10.7--12.7 & 2.9--4.3\,GB \\
\bottomrule
\end{tabular}
\caption{Head-ranking cost across six models on one NVIDIA A100 PCIe 40\,GB GPU. Time excludes the shared forward pass, and ratios use the approximation in the same setting as the reference. 
Extra memory excludes the model and shared forward pass and covers all rows in batched runs.}
\label{tab:app-cost}
\end{table*}

\section{Baselines and Implementation}
\label{supp:baselines}

All baselines explain the same token at the same predicting row as \wish and return a map on the visual-token grid,
using the same generated captions, word grouping and targets. For perturbation targets with several tokens, we
normalize the map of each token to unit sum and add these maps (Sec.~\ref{app:targets}). We then min--max normalize every map over the visual
tokens for localization and display. We use the following baselines:

\noindent\textbf{TAM~\cite{li-token-activation-map-2025}.} We use the authors' official implementation.

\noindent\textbf{DEX-AR~\cite{bousselham-dexar-2026}.} We use the authors' official implementation.

\noindent\textbf{Attention.} This baseline uses only the attention weights of the row that predicts $y_t$ and needs
no gradient. For each visual token, the map averages the weight over the heads of a layer and sums these averages over
the layers:
\begin{equation}
M^{\mathrm{Att}}_{t,j}=\frac{1}{H}\sum_{\ell=1}^{L}\sum_{h=1}^{H}A_j^{\ell,h},\qquad j\in\mathcal I,
\label{eq:app-base-attention}
\end{equation}
where $M^{\mathrm{Att}}_{t,j}$ is the map value at visual token $j$ and $A_j^{\ell,h}$ is the attention weight of head
$(\ell,h)$ from the predicting row to that token (Eq.~\eqref{eq:app-attention}).

\noindent\textbf{Attn$\times$Grad~\cite{chefer-beyond-attention-2021}.} This baseline weights each attention weight by
its gradient. Each layer is differentiated through its own layer logit (Eq.~\eqref{eq:app-logit-lens}), which gives
the attention-gradient $G_j^{\ell,h}$ of every head and visual token. The map keeps the positive part of the product
of weight and gradient and sums it over all heads and layers:
\begin{equation}
M^{\mathrm{AG}}_{t,j}=\sum_{\ell=1}^{L}\sum_{h=1}^{H}\bigl\lfloor A_j^{\ell,h}G_j^{\ell,h}\bigr\rfloor_+,\qquad
j\in\mathcal I,
\label{eq:app-base-attngrad}
\end{equation}
where $M^{\mathrm{AG}}_{t,j}$ is the map value at visual token $j$. 

\noindent\textbf{Grad-CAM~\cite{selvaraju-gradcam-2017}.} This baseline works on the hidden states of the visual
tokens instead of the attention following DEX-AR's implementation~\cite{bousselham-dexar-2026}. We use the hidden states that enter the first decoder layer, which are the visual
tokens after the projector. The model's output logit of $y_t$ is the layer logit of the last layer,
$\hat o^{L}(y_t)$. For each visual token, the map multiplies every feature of the hidden state by the gradient of this
logit with respect to that feature, keeps the positive part and sums over the features:
\begin{equation}
M^{\mathrm{GC}}_{t,j}=\sum_{c=1}^{d}\Bigl\lfloor\frac{\partial\hat o^{L}(y_t)}{\partial z^{0}_{j,c}}\,z^{0}_{j,c}\Bigr\rfloor_+,
\qquad j\in\mathcal I,
\label{eq:app-base-gradcam}
\end{equation}
where $M^{\mathrm{GC}}_{t,j}$ is the map value at visual token $j$ and $z^{0}_{j,c}$ is feature $c$ of the hidden
state $\mathbf z_j^{0}\in\mathbb R^d$ of visual token $j$ at the input of the first decoder layer.

\noindent\textbf{Chefer-CAM~\cite{chefer-generic-attention-2021}.} This baseline propagates relevance through all
attention layers following DEX-AR's implementation~\cite{bousselham-dexar-2026}. At each layer, it weights every attention weight by its gradient with respect to the output logit,
sums over the heads and keeps the positive part:
\begin{equation}
\bar{\mathbf A}^{\ell}=\Bigl\lfloor\sum_{h=1}^{H}\mathbf A^{\ell,h}\odot
\frac{\partial\hat o^{L}(y_t)}{\partial\mathbf A^{\ell,h}}\Bigr\rfloor_+,
\label{eq:app-base-chefer-layer}
\end{equation}
where $\bar{\mathbf A}^{\ell}$ is the relevance-weighted attention of layer $\ell$, $\mathbf A^{\ell,h}$ is the full
attention matrix of head $(\ell,h)$ (Eq.~\eqref{eq:app-attention-matrix}) and $\odot$ is the elementwise product. The
relevance matrix starts from the identity and is updated layer by layer:
\begin{equation}
\mathbf R^{0}=\mathbf I,\qquad \mathbf R^{\ell}=\mathbf R^{\ell-1}+\bar{\mathbf A}^{\ell}\mathbf R^{\ell-1},
\label{eq:app-base-chefer-prop}
\end{equation}
where $\mathbf R^{\ell}$ is the relevance after layer $\ell$. The map is the row of the last relevance matrix that
belongs to the predicting position, read at the visual tokens:
\begin{equation}
M^{\mathrm{CC}}_{t,j}=\mathbf R^{L}_{-1,j},\qquad j\in\mathcal I,
\label{eq:app-base-chefer-map}
\end{equation}
where $M^{\mathrm{CC}}_{t,j}$ is the map value at visual token $j$. Only this row is needed. Since
$\mathbf R^{L}=(\mathbf I+\bar{\mathbf A}^{L})\cdots(\mathbf I+\bar{\mathbf A}^{1})$, we multiply a row vector by these
factors from the last layer to the first. This gives the same values without forming the full matrices.

\section{Comparison with EAGLE}
\label{supp:eagle}

We compare \wish with EAGLE~\cite{chen-eagle-2026} in terms of localization, perturbation, and computational cost.
EAGLE uses a greedy submodular search to select superpixels that
increase the target word's probability when retained and decrease it
when removed.

\subsection{Localization and Perturbation}
\label{app:eagle-maps}

We evaluate both methods on object words generated by each of the
six models on 150 PASCAL VOC validation images. We score each word
at its generated position with the caption prefix fixed. For EAGLE,
we use the authors' implementation with 64 SLICO superpixels and
$\lambda_1=\lambda_2=1$, adapting only the masked-image scoring call
to use the same preprocessing and target positions as \wish. 
For localization, we score EAGLE's maps at image resolution and upsample \wish's maps as in Tab.~\ref{tab:localization} of the main paper.
For perturbation, we average EAGLE's map within each visual-token cell and follow Tab.~\ref{tab:perturbation} of the main paper.

In Tab.~\ref{tab:app-eagle}, EAGLE achieves the better insertion score
on all six models and the higher Pointing Game score on five,
indicating that its maps identify informative regions and often peak
inside the object. \wish achieves the higher Obj-IoU and Soft-IoU on all six
models, showing closer agreement with the full object masks. It also
assigns a larger fraction of saliency to the object than EAGLE on
every model, as measured by EPG. In the deletion test, \wish
outperforms EAGLE on five models, with GLM-4.6V-Flash as the exception.
The test replaces pixels with the dataset mean, whereas EAGLE uses
black masking in its search.

\begin{table}[t]
\centering
\small
\setlength{\tabcolsep}{3.5pt}
\begin{tabular}{@{}lcccccc@{}}
\toprule
Method & PG & Ins.\ $\downarrow$ & Del. $\uparrow$ & Obj-IoU & Soft-IoU & EPG \\
\midrule
\multicolumn{7}{@{}l}{\textit{LLaVA-1.5-7B} (102 words)} \\
EAGLE & \textbf{85.3} & \textbf{0.18} & 1.45 & 20.0 & 15.2 & 20.6 \\
\wish & 75.5 & 0.33 & \textbf{2.99} & \textbf{45.8} & \textbf{26.2} & \textbf{44.5} \\
\midrule
\multicolumn{7}{@{}l}{\textit{BakLLaVA} (98 words)} \\
EAGLE & 77.6 & \textbf{0.33} & 2.01 & 16.6 & 14.8 & 20.5 \\
\wish & \textbf{82.7} & 0.48 & \textbf{3.85} & \textbf{46.1} & \textbf{26.4} & \textbf{42.9} \\
\midrule
\multicolumn{7}{@{}l}{\textit{Qwen3-VL-8B} (91 words)} \\
EAGLE & \textbf{68.1} & \textbf{0.33} & 1.57 & 13.5 & 12.9 & 19.7 \\
\wish & 38.5 & 0.55 & \textbf{1.76} & \textbf{29.1} & \textbf{18.8} & \textbf{29.1} \\
\midrule
\multicolumn{7}{@{}l}{\textit{GLM-4.6V-Flash} (111 words)} \\
EAGLE & \textbf{73.0} & \textbf{0.46} & \textbf{1.27} & 14.0 & 12.0 & 16.4 \\
\wish & 21.6 & 0.68 & 1.13 & \textbf{20.4} & \textbf{13.2} & \textbf{23.6} \\
\midrule
\multicolumn{7}{@{}l}{\textit{LLaVA-OV-1.5-8B} (112 words)} \\
EAGLE & \textbf{75.9} & \textbf{0.36} & 1.36 & 16.7 & 13.9 & 19.3 \\
\wish & 33.9 & 0.46 & \textbf{2.31} & \textbf{30.1} & \textbf{18.9} & \textbf{29.7} \\
\midrule
\multicolumn{7}{@{}l}{\textit{LLaVA-OV-2-8B} (107 words)} \\
EAGLE & \textbf{76.6} & \textbf{0.37} & 1.18 & 15.7 & 12.9 & 17.7 \\
\wish & 33.6 & 0.52 & \textbf{2.23} & \textbf{30.4} & \textbf{19.4} & \textbf{29.5} \\
\bottomrule
\end{tabular}
\caption{Localization and perturbation on object words from 150 PASCAL VOC validation images. Pointing Game (PG):
share of words whose map peak lies inside the object mask. Other metrics as in Tabs.~\ref{tab:perturbation} and~\ref{tab:localization} of the main paper.}
\label{tab:app-eagle}
\end{table}

\subsection{Inference Time and Memory}
\label{app:cost-methods}

We measure the time and peak GPU memory that each method requires to produce an explanation map on NVIDIA RTX PRO 6000 Blackwell GPUs, using one GPU per run and the default precision of each model.
The time includes the forward pass and excludes model loading.
All methods use eager attention and compute their maps on the GPU.
For TAM, DEX-AR and \wish, we report the median time per word on 30 PASCAL VOC validation images per model with teacher-forced captions.
For EAGLE, we time the first token of five object words per model and exclude the superpixel segmentation.

We compare the cost of \wish with that of EAGLE and report TAM and DEX-AR as references.
We time TAM, DEX-AR and \wish both one word at a time and batched over the words of a caption while EAGLE searches each word separately and has no batched setting, so we time it only one word at a time.
 In Tab.~\ref{tab:app-cost-time}, TAM is the fastest method on all six models and in both settings, followed by \wish
and DEX-AR. TAM needs no backward pass. \wish takes 1.4--1.5 times as long as TAM, and DEX-AR takes 1.3--2.1 times
as long as \wish. TAM and \wish need at most 0.9\,GB of GPU memory beyond the model, while DEX-AR needs
2.8--12.6\,GB. EAGLE runs $4{,}069$--$4{,}120$ forward passes per word and takes $2{,}600$--$3{,}000$ times as long as \wish run one word at a time.

\begin{table*}[t]
\centering
\footnotesize
\setlength{\tabcolsep}{2pt}
\begin{tabular}{@{}lccccccc@{\hspace{14pt}}cccccccc@{}}
\toprule
& \multicolumn{7}{c}{Time per word (ms)} & \multicolumn{8}{c}{Peak GPU memory (GB)} \\
\cmidrule(lr){2-8}\cmidrule(l){9-16}
& \multicolumn{3}{c}{One word} & \multicolumn{3}{c}{Caption-batched} & & & \multicolumn{3}{c}{One word} & \multicolumn{3}{c}{Caption-batched} & \\
\cmidrule(lr){2-4}\cmidrule(lr){5-7}\cmidrule(lr){10-12}\cmidrule(lr){13-15}
Model & TAM & DEX-AR & \wish & TAM & DEX-AR & \wish & EAGLE & Model alone & TAM & DEX-AR & \wish & TAM & DEX-AR & \wish & EAGLE \\
\midrule
LLaVA-1.5-7B & 48.2 & 84.7 & 66.8 & 3.5 & 6.8 & 4.9 & 198,000 & 13.2 & 13.5 & 24.9 & 14.0 & 13.5 & 19.3 & 14.1 & 13.8 \\
BakLLaVA & 50.2 & 90.8 & 70.7 & 3.9 & 7.7 & 5.5 & 203,000 & 14.1 & 14.4 & 26.7 & 14.4 & 14.4 & 20.7 & 14.5 & 14.7 \\
Qwen3-VL-8B & 47.0 & 123.2 & 66.5 & 1.7 & 4.7 & 2.4 & 188,000 & 16.3 & 16.7 & 21.4 & 16.5 & 16.8 & 19.1 & 16.7 & 17.0 \\
GLM-4.6V-Flash & 57.4 & 151.3 & 87.1 & 1.0 & 2.6 & 1.5 & 226,000 & 19.2 & 19.8 & 29.4 & 19.4 & 19.8 & 24.5 & 19.9 & 19.8 \\
LLaVA-OV-1.5-8B & 46.6 & 122.1 & 67.1 & 3.3 & 9.6 & 4.6 & 185,000 & 15.9 & 16.4 & 22.2 & 16.2 & 16.4 & 19.2 & 16.3 & 16.6 \\
LLaVA-OV-2-8B & 43.1 & 120.7 & 63.6 & 3.1 & 9.0 & 4.4 & 175,000 & 15.9 & 16.4 & 22.2 & 16.2 & 16.4 & 19.2 & 16.3 & 16.6 \\
\bottomrule
\end{tabular}
\caption{Map-generation time per word and peak allocated GPU memory of \wish and EAGLE, with TAM and DEX-AR as references, on one NVIDIA RTX PRO 6000 Blackwell GPU. 
TAM, DEX-AR and \wish run one word at a time and batched over the words of a caption. EAGLE runs one word at a time only. Model alone: the memory of the model without an explanation method.}
\label{tab:app-cost-time}
\end{table*}

\section{Evaluation Protocols}
\label{app:protocols}

\subsection{Evaluation Setup}
\label{app:targets}
\paragraph{Models.} We evaluate LLaVA-1.5-7B~\cite{liu-llava15-2024}, BakLLaVA~\cite{skunkworks-bakllava-2023}, Qwen3-VL-8B~\cite{qwen3-vl-2025}, GLM-4.6V-Flash~\cite{vteam-glm45v-2025}, LLaVA-OneVision-1.5-8B~\cite{an-llava-onevision-15-2025} and LLaVA-OneVision-2-8B~\cite{an-llava-onevision-2-2026}.
We write LLaVA-OneVision as LLaVA-OV, or OV where space is short, and drop the model size in some tables.

\paragraph{Free generation.} Each model captions the image with greedy decoding, the prompt ``Write a one-sentence
caption for this image:'' and at most 64 new tokens. Every method explains the generated tokens at the positions
where they were generated. All methods use the same caption.

\paragraph{Teacher forcing.} For ImageNet, the target is the class name, the first WordNet synonym of the class,
with the prompt ``Classify the image.''. For VQAv2, it is the most frequent human answer, with the prompt ``Answer the question using a single word or phrase.''. 
Each prompt is wrapped in the model's own chat template. 
The target is appended to the prompt and explained at its own positions.

\paragraph{Object words.} We group tokens into words with TAM's word grouping~\cite{li-token-activation-map-2025}.
Following TAM's matching rule, a caption word is an object word when its WordNet lemma matches the lemma of a word in
the name of an annotated class present in the image. Thus ``table'' matches \emph{dining table} and ``plant'' matches
\emph{potted plant}. The word's mask is the segmentation mask of that class. For PASCAL VOC, the class names are the
20 category names in PASCAL VOC's own spellings (\emph{aeroplane}, \emph{diningtable}, \emph{pottedplant},
\emph{tvmonitor}). For COCO, they are the names in TAM's category dictionary. On both datasets, object words use no
synonyms, so words such as ``man'' or ``people'' do not count as \emph{person}.

\paragraph{Words with several sub-tokens.} Every sub-token receives its own map. Maps are never merged across the
sub-tokens of a word for localization or display. Following TAM's protocol, a word's Obj-IoU is the highest Obj-IoU
among its sub-tokens' maps. Soft-IoU and EPG are measured on that same sub-token, and the qualitative figures show its
map. The perturbation targets with several tokens are ImageNet class names, VQAv2 answers and COCO captions. For them,
each token's map is normalized to unit sum and the maps are summed into one map. For function words, all quantities use first sub-tokens. The scores use the function word's map, Del.\ reads the
function word's log-probability and FWP reads the log-probability of its noun.

\subsection{Datasets}
\label{app:datasets}

\noindent\textbf{PASCAL VOC 2012~\cite{everingham-voc-2010}.} All 1{,}449 segmentation validation images are used
for every PASCAL VOC evaluation, with two exceptions. The fixed head budget is tuned on the training images (Sec.~\ref{app:budget}) and the comparison with EAGLE uses 150 validation images (Sec.~\ref{app:eagle-maps}).

\noindent\textbf{COCO Caption~\cite{chen-coco-captions-2015}.} The 5{,}000 images of COCO 2014 minival as released by
TAM~\cite{li-token-activation-map-2025}, with the object masks of each class from that release, derived from the COCO
annotations~\cite{lin-coco-2014}. 
A fixed subset of 1{,}000 of these images, drawn with seed 0, is used for perturbation and for the knockout of Sec.~\ref{sec:head_selection} of the main paper.

\noindent\textbf{ImageNet and VQAv2.} Following DEX-AR~\cite{bousselham-dexar-2026}, 5{,}000 ImageNet~\cite{deng-imagenet-2009}
validation images (five per class) and 1{,}000 VQAv2~\cite{goyal-vqav2-2017} validation image--question pairs, as fixed
lists shared by every method and model.

\subsection{Attention Knockout Evaluation Protocol}
\label{app:knockout-protocol}

\paragraph{Intervention.} Every head in the selected set is blocked from the whole image ($R=\mathcal I$) at the row
that predicts $y_t$, and the remaining keys are renormalized as in Eq.~\eqref{eq:app-renorm}. All selected heads are
blocked jointly in one forward pass with the caption prefix fixed, and we read the drop $\delta_t(S_t,R)$ in the final
log-probability of the target word (Sec.~\ref{sec:head_selection}).

\paragraph{Reported quantity.} For each model, the mean of $\delta_t(S,R)$ over words for every ranking, divided by
the mean for \wish on the same words (so \wish{} $=1$). Significance per model comes from a 95\% image bootstrap of
the paired difference to \wish.

\subsection{Head Rankings}
\label{app:head-rankings}

\paragraph{Baselines.} In Fig.~\ref{fig:knockout}, every ranking scores all $LH$ heads at the row that predicts $y_t$ and keeps its
$|S_t|$ highest-scoring heads. Here $S_t$ is \wish's selected set for the same word, so all sets have the same size
per word. Two rankings use $\mathcal I_{\bar M_t}\subset\mathcal I$, the visual tokens covered by the word's
ground-truth segmentation mask $\bar M_t$:
\begin{itemize}
\item \textbf{\emph{\wish}}: the knee set $S_t$ of $s_t^{\ell,h}$ (Eq.~\eqref{eq:wish-selected-heads}).
\item \textbf{\emph{DEX-AR weight}}: DEX-AR's head weight~\cite{bousselham-dexar-2026},
      \[s_{\text{DEX-AR}}^{\ell,h}=\bigl\lfloor\max_{j\in\mathcal I}P_j^{\ell,h}-\max_{k\in\mathcal K\setminus\mathcal I}P_k^{\ell,h}\bigr\rfloor_+,\]
      over heads with $s_{\text{DEX-AR}}^{\ell,h}>0$.
\item \textbf{\emph{Attention mass}}: the head's attention to the image,
      \[s_{\text{AM}}^{\ell,h}=\sum_{j\in\mathcal I}A_j^{\ell,h}.\]
\item \textbf{\emph{Mask fit}} (uses labels): enrichment of the head's gradient map inside the object,
      \[s_{\text{MF}}^{\ell,h}=\dfrac{\sum_{j\in\mathcal I_{\bar M_t}}P_j^{\ell,h}\,/\sum_{j\in\mathcal I}P_j^{\ell,h}}{|\mathcal I_{\bar M_t}|/|\mathcal I|}.\]
\item \textbf{\emph{Attention on mask}} (uses labels): the head's attention to the object,
      $s_{\text{AoM}}^{\ell,h}=\sum_{j\in\mathcal I_{\bar M_t}}A_j^{\ell,h}$.
\item \textbf{\emph{Spatial coherence}}: Moran's $I$~\cite{moran1950notes} of the head's gradient map on the visual-token grid,
      \[s_{\text{SC}}^{\ell,h}=\dfrac{|\mathcal I|}{|E|}\,\dfrac{\sum_{(j,j')\in E}z_jz_{j'}}{\sum_{j\in\mathcal I}z_j^2},\]
      where
      $z_j=P_j^{\ell,h}-\frac{1}{|\mathcal I|}\sum_{j'\in\mathcal I}P_{j'}^{\ell,h}$ and $E$ is the set of horizontally or vertically adjacent cell pairs.
\item \textbf{\emph{Random}}: $|S_t|$ heads drawn uniformly, seeded per word.
\end{itemize}
Ties are broken by the sort order. Mask fit and attention on mask serve as references that know the object and do not compete with the methods that use no labels.

\paragraph{Downstream Effect of Head Selection on the Maps.}
\label{app:selection-downstream}

The knockout (Sec.~\ref{sec:head_selection}) measures how much the selected heads matter for the predicted word. Here
we test whether the heads that each ranking selects also give good maps. For each ranking we keep the 64 highest-ranked heads and
build the map as \wish does (Sec.~\ref{app:aggregation}). Each head's map is filtered with a $3\times3$ median, and the
maps are combined by the weighted median with linear rank weights. The rank comes from the ranking itself. Random heads
are weighted by their rank under $s_t^{\ell,h}$, as in the random row of Tab.~\ref{tab:ablations} of the main paper. We score the maps on
PASCAL VOC with the metrics of Tab.~\ref{tab:ablations} of the main paper (Tab.~\ref{tab:app-crit-maps}). Among the rankings that use no
labels, \wish's heads give the best maps on all four metrics. Mask fit selects heads whose gradient maps already lie
inside the annotated object. Its maps localize better than \wish's, but they delete less in deletion perturbation tests ($2.07$ vs.\ $2.19$).
Attention on mask localizes no better than random heads ($16.7$ vs.\ $17.5$ Obj-IoU). Where a head attends therefore
says little about where its evidence lies.

\begin{table}[t]
\centering
\small
\setlength{\tabcolsep}{3pt}
\begin{tabular}{@{}lcccc@{}}
\toprule
Ranking (64 heads) & Obj-IoU$\uparrow$ & Soft-IoU$\uparrow$ & EPG$\uparrow$ & Del.$\uparrow$ \\
\midrule
\wish & \textbf{36.7} & \textbf{22.1} & \textbf{38.4} & \textbf{2.19} \\
DEX-AR weight & 30.3 & 18.8 & 33.6 & 1.89 \\
Attention mass & 18.8 & 12.4 & 29.0 & 1.45 \\
Spatial coherence & 29.8 & 21.3 & 33.3 & 1.80 \\
Random & 17.5 & 12.6 & 25.4 & 1.03 \\
\midrule
Mask fit (labels) & 41.3 & 26.8 & 75.6 & 2.07 \\
Attention on mask (labels) & 16.7 & 11.0 & 27.3 & 1.31 \\
\bottomrule
\end{tabular}
\caption{Maps built with \wish's map construction from the 64 highest-ranked heads of each head scoring method. PASCAL VOC,
six-model means, metrics as in Tab.~\ref{tab:ablations} of the main paper. Del.\ uses the 896--1{,}106 words per model for which every
ranking gives a map. Labels: the ranking uses the annotated object mask. Bold: best ranking without labels.}
\label{tab:app-crit-maps}
\end{table}

\section{Function-Word Grounding}
\label{app:function-words}

\paragraph{Word extraction.}
We identify function words whose syntactic head is a noun naming an annotated object.
On PASCAL VOC, a noun can also name a class through a fixed synonym list, unlike the object words of
Sec.~\ref{app:targets}. The list maps ``man'', ``woman'', ``men'', ``women'', ``people'', ``boy'', ``girl'', ``child'',
``children'', ``kid'', ``player'', ``rider'', ``lady'', ``guy'', ``skier'' and ``surfer'' to \emph{person}. It maps
``plane'', ``aircraft'', ``jet'' and ``airliner'' to \emph{aeroplane}, ``bike'' to \emph{bicycle}, ``ship'',
``sailboat'' and ``yacht'' to \emph{boat}, ``motorcycle'' to \emph{motorbike}, ``couch'' to \emph{sofa}, ``locomotive''
to \emph{train}, and ``television'', ``tv'', ``screen'' and ``monitor'' to \emph{tvmonitor}.
We parse each caption with Stanza's English Universal Dependencies (UD) pipeline~\cite{nivre-ud-2020,qi-stanza-2020}, and retain four dependency relations: determiners (\texttt{det}), prepositions and other case markers (\texttt{case}), coordinating conjunctions (\texttt{cc}), and numerals (\texttt{nummod}).
Auxiliaries, subordinating conjunctions, and particles typically attach to predicates or clauses. We exclude possessive pronouns because they rarely modify nouns naming annotated objects (0--12 instances per model).
Pronouns such as ``it'' and ``them'' typically refer to objects already mentioned in the caption.

\paragraph{Evaluated nouns.} For Del., FWP-AUC and FWS (Tab.~\ref{tab:function_words} of the main paper), the function word must precede
its noun, and the noun must be the first mention of its object class in the caption. The first rule removes four
possessive markers ``'s'' across all models, and the second removes 154--369 candidates per model. We then keep one
function word per object, the first that introduces it, and evaluate captions in which at least two objects are
introduced this way. This gives 2{,}984 nouns, 375--544 per model (Tab.~\ref{tab:function_words_models}). Each noun
is assigned to the category of its function word. Each model's result averages the four category means, so that each category counts equally.

\paragraph{Empty maps.} Heatmaps for function words are sometimes zero everywhere for TAM. In Tab.~\ref{tab:function_words} of the main paper, an empty map removes no cells, so its drops (Del., FWP-AUC) and its FWS are zero.

\paragraph{Results per model.} Tab.~\ref{tab:function_words_models} reports Del., FWP-AUC and FWS of Tab.~\ref{tab:function_words} of the main paper for each model. \wish leads on five of six models on each metric. Grad-CAM leads Del.\ on GLM-4.6V, Chefer-CAM leads FWP-AUC on LLaVA-1.5 and TAM leads FWS on LLaVA-OV-2.

\begin{table}[t]
  \centering
  \footnotesize
  \setlength{\tabcolsep}{0.7pt}
  \begin{tabular}{@{}lcccccc@{}}
    \toprule
    Method & Qwen3-VL & LLaVA-1.5 & GLM-4.6V & BakLLaVA & OV-1.5 & OV-2 \\
    Nouns & 375 & 508 & 537 & 499 & 544 & 521 \\
    \midrule
    \multicolumn{7}{@{}l}{\emph{Del.$\uparrow$}} \\
    Attention & 2.28 & 0.83 & 1.36 & 1.35 & 0.92 & 0.69 \\
    Attn$\times$Grad & 2.32 & 0.84 & 1.38 & 1.35 & 1.01 & 0.68 \\
    Grad-CAM & 2.43 & 0.80 & \textbf{1.84} & \underline{1.38} & \underline{1.06} & 0.67 \\
    Chefer-CAM & 2.47 & 0.70 & \underline{1.82} & 1.37 & 0.92 & 0.71 \\
    DEX-AR & \underline{2.51} & \underline{0.94} & 1.69 & 1.13 & 0.99 & 0.67 \\
    TAM & 1.78 & 0.48 & 1.27 & 0.80 & 0.87 & \underline{0.95} \\
    \wish & \textbf{2.98} & \textbf{1.01} & 1.73 & \textbf{1.40} & \textbf{1.26} & \textbf{1.05} \\
    \midrule
    \multicolumn{7}{@{}l}{\emph{FWP-AUC$\uparrow$}} \\
    Attention & 0.46 & 0.26 & 0.34 & \underline{0.33} & 0.29 & 0.36 \\
    Attn$\times$Grad & 0.43 & \underline{0.31} & 0.40 & 0.29 & \underline{0.33} & 0.32 \\
    Grad-CAM & 0.51 & 0.25 & \underline{0.60} & 0.25 & 0.25 & 0.24 \\
    Chefer-CAM & \underline{0.53} & \textbf{0.32} & 0.52 & 0.25 & 0.12 & 0.17 \\
    DEX-AR & 0.17 & 0.06 & 0.09 & 0.05 & 0.18 & 0.19 \\
    TAM & 0.25 & $-$0.05 & 0.27 & $-$0.13 & 0.12 & \underline{0.50} \\
    \wish & \textbf{0.55} & 0.19 & \textbf{0.73} & \textbf{0.54} & \textbf{0.50} & \textbf{0.59} \\
    \midrule
    \multicolumn{7}{@{}l}{\emph{FWS$\uparrow$ ($\times 100$)}} \\
    Attention & 0.50 & 0.36 & 1.37 & 0.53 & 0.91 & 0.57 \\
    Attn$\times$Grad & 0.79 & 0.82 & 1.57 & 0.86 & \underline{1.91} & 1.35 \\
    Grad-CAM & 0.93 & 0.71 & \underline{1.69} & 0.38 & 0.90 & 0.68 \\
    Chefer-CAM & 0.33 & \underline{1.61} & 1.41 & \underline{1.28} & 0.36 & 0.45 \\
    DEX-AR & 1.01 & 1.37 & 0.16 & 0.58 & 0.62 & 0.86 \\
    TAM & \underline{2.50} & $-$0.62 & 0.98 & $-$0.70 & 1.04 & \textbf{3.08} \\
    \wish & \textbf{2.71} & \textbf{2.95} & \textbf{2.03} & \textbf{2.56} & \textbf{3.40} & \underline{2.45} \\
    \bottomrule
  \end{tabular}
  \caption{Per-model results of Tab.~\ref{tab:function_words} of the main paper: positive deletion with the function word as target (Del.), FWP-AUC and FWS, each averaged over the function-word categories of a model. Best per model in bold, second best underlined.}
  \label{tab:function_words_models}
\end{table}

\subsection{Visualizing FWS and FWP-AUC}
\label{app:fw-examples}

\begin{figure*}[t]
\centering
\includegraphics[width=\textwidth]{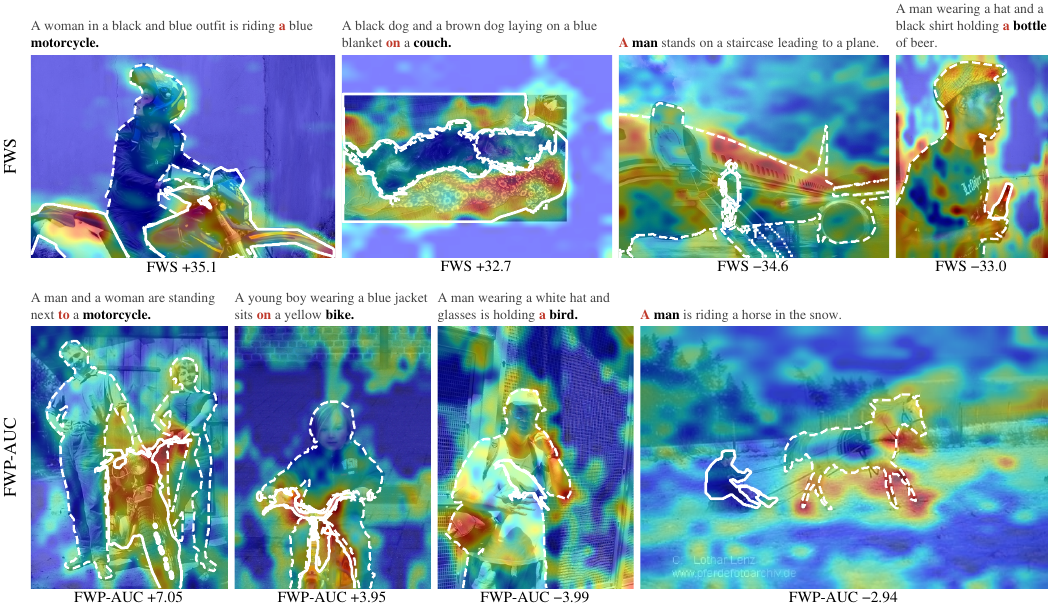}
\caption{\wish's maps of function words with high and low FWS (top, $\times 100$) and FWP-AUC (bottom) on PASCAL VOC,
from LLaVA-1.5-7B (first panel: BakLLaVA). Red: function word. Bold: the noun it introduces. Solid and dashed contours:
masks of the introduced object and of the other introduced object.}
\label{fig:app-fw-examples}
\end{figure*}

Fig.~\ref{fig:app-fw-examples} shows \wish's maps of function words with high and low scores for LLaVA-1.5-7B and
BakLLaVA on PASCAL VOC. Each panel shows the map of one function word, the mask of the object it introduces (solid
contour) and the mask of the other object introduced in the caption (dashed contour). A high FWS means that the map
overlaps the introduced object more than the other object. A high FWP-AUC means that replacing the cells that the map
ranks highest lowers the noun's log-probability more than replacing the cells of the other function word's map. In the
high examples of both rows, the map lies on the introduced object. In the low examples, it lies on the other object.

\subsection{Function Words in Other Languages}
\label{app:fw-languages}

\begin{figure*}[t]
\centering
\includegraphics[width=\textwidth]{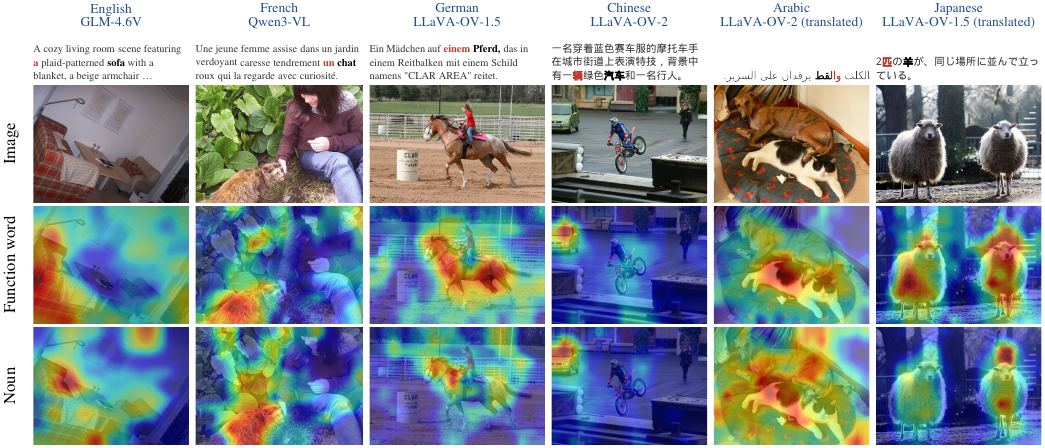}
\caption{\wish's maps of a function word and of its noun in English and five other languages, each for the first
token of the word. Above each column: language, model and the model's caption, with the function word in red and the
noun in bold. Translated: the model's own translation of its English caption, explained with teacher forcing.}
\label{fig:app-fw-languages}
\end{figure*}

Function words occur in all languages, but their grammatical realization varies. French and German introduce nouns
with articles, and Arabic marks definiteness with the article al-, which is attached to the noun as a prefix. Chinese
and Japanese have no articles. Instead, a classifier accompanies a noun when it is counted, and particles mark
grammatical relations, in Japanese after the noun they mark. To illustrate that \wish is not restricted to English,
Fig.~\ref{fig:app-fw-languages} presents qualitative examples in English and five other languages. For each example, we show the map of a function word alongside the map of its associated noun, computed on captions that the models generated.

\section{Additional Results}
\label{app:additional}

\subsection{Ablation per Model}
\label{app:ablation-models}

Tab.~\ref{tab:app-ablation-models} reports the ablation of Tab.~\ref{tab:ablations} of the main paper per model. 
The adaptive knee matches the fixed Top-64 budget on every model without choosing $K$. 
The differences are at most $0.01$ positive-deletion AUC, $0.6$ Obj-IoU, $0.3$ Soft-IoU and $0.6$ EPG. 
The best fixed budget on the training images also differs between models. On GLM-4.6V-Flash, it is $K=128$ rather than $64$ (Fig.~\ref{fig:app-budget}). A fixed budget therefore has to be tuned, and the tuned value is not the best one for every model. 
The knee instead sets the number of heads for each word.

\begin{table*}[t]
\centering
\footnotesize
\renewcommand{\arraystretch}{1.06}
\setlength{\tabcolsep}{0.3pt}
\begin{tabular*}{\textwidth}{@{\extracolsep{\fill}}clcccccc@{\hspace{5pt}}cccccc@{\hspace{5pt}}cccccc@{\hspace{5pt}}cccccc@{}}
\toprule
 & & \multicolumn{6}{c@{\hspace{5pt}}}{Del.$\uparrow$} & \multicolumn{6}{c@{\hspace{5pt}}}{Obj-IoU$\uparrow$} & \multicolumn{6}{c@{\hspace{5pt}}}{Soft-IoU$\uparrow$} & \multicolumn{6}{c}{EPG$\uparrow$} \\
\cmidrule(lr){3-8}\cmidrule(lr){9-14}\cmidrule(lr){15-20}\cmidrule(l){21-26}
 & Variant & \rotatebox{90}{Qwen3-VL} & \rotatebox{90}{LLaVA-1.5} & \rotatebox{90}{GLM-4.6V} & \rotatebox{90}{BakLLaVA} & \rotatebox{90}{OV-1.5} & \rotatebox{90}{OV-2} & \rotatebox{90}{Qwen3-VL} & \rotatebox{90}{LLaVA-1.5} & \rotatebox{90}{GLM-4.6V} & \rotatebox{90}{BakLLaVA} & \rotatebox{90}{OV-1.5} & \rotatebox{90}{OV-2} & \rotatebox{90}{Qwen3-VL} & \rotatebox{90}{LLaVA-1.5} & \rotatebox{90}{GLM-4.6V} & \rotatebox{90}{BakLLaVA} & \rotatebox{90}{OV-1.5} & \rotatebox{90}{OV-2} & \rotatebox{90}{Qwen3-VL} & \rotatebox{90}{LLaVA-1.5} & \rotatebox{90}{GLM-4.6V} & \rotatebox{90}{BakLLaVA} & \rotatebox{90}{OV-1.5} & \rotatebox{90}{OV-2} \\
\midrule
\multirow{4}{*}{\rotatebox[origin=c]{90}{Selection}}
 & Random (knee-sized) & 1.06 & 1.05 & 0.66 & 1.74 & 0.96 & 0.85 & 16.1 & 20.6 & 13.3 & 21.6 & 16.8 & 16.2 & 12.0 & 13.8 & 9.9 & 14.6 & 12.7 & 12.0 & 24.1 & 28.7 & 21.9 & 29.6 & 23.8 & 23.9 \\
 & All heads & 1.68 & \underline{1.98} & \underline{0.90} & 2.86 & \underline{1.58} & 1.55 & 24.0 & 33.0 & 18.0 & \underline{34.4} & 24.7 & 24.6 & \underline{18.2} & \underline{21.1} & 14.1 & 22.6 & \underline{18.6} & 18.4 & 27.0 & 31.1 & 22.9 & 33.0 & 25.6 & 26.0 \\
 & $s_t^{\ell,h}$: Top-64 & \textbf{2.00} & \textbf{2.71} & \textbf{1.32} & \textbf{3.37} & \textbf{1.98} & \underline{1.90} & \textbf{30.1} & \textbf{51.0} & \textbf{23.5} & \textbf{50.1} & \underline{33.0} & \underline{32.3} & \textbf{19.9} & \textbf{28.2} & \textbf{15.0} & \underline{28.6} & \textbf{20.4} & \textbf{20.5} & \underline{32.6} & \underline{50.4} & \textbf{29.6} & \textbf{50.5} & \underline{33.6} & \underline{33.6} \\
 & \textit{(Ours)} $s_t^{\ell,h}$: Knee & \underline{1.99} & \textbf{2.71} & \textbf{1.32} & \underline{3.36} & \textbf{1.98} & \textbf{1.91} & \underline{30.0} & \underline{50.9} & \underline{22.9} & \textbf{50.1} & \textbf{33.2} & \textbf{32.4} & \textbf{19.9} & \textbf{28.2} & \underline{14.7} & \textbf{28.7} & \textbf{20.4} & \underline{20.4} & \textbf{32.7} & \textbf{51.0} & \underline{29.4} & \underline{50.4} & \textbf{34.0} & \textbf{33.8} \\
\midrule
\multirow{4}{*}{\rotatebox[origin=c]{90}{Filtering}}
 & None & 1.72 & 2.62 & 1.13 & 3.27 & \underline{1.81} & 1.65 & 19.4 & 44.2 & 16.2 & 44.6 & 24.2 & 23.5 & 11.7 & 23.0 & 10.6 & 24.1 & 14.2 & 12.9 & \underline{30.1} & \underline{48.5} & \underline{27.6} & \underline{47.8} & \underline{31.5} & \underline{31.7} \\
 & $3{\times}3$ mean & 1.79 & \underline{2.71} & \textbf{1.36} & \textbf{3.37} & 1.75 & 1.76 & \underline{23.3} & 49.3 & \underline{22.2} & 49.5 & \underline{28.5} & \underline{28.3} & \underline{16.9} & \underline{26.8} & \textbf{15.9} & \underline{26.4} & \underline{19.0} & \underline{18.7} & 26.5 & 40.1 & 25.2 & 38.2 & 28.5 & 28.3 \\
 & $3{\times}3$ Gaussian ($\sigma=1$) & \underline{1.83} & \textbf{2.72} & \textbf{1.36} & \textbf{3.37} & 1.75 & \underline{1.78} & \underline{23.3} & \underline{49.6} & 21.8 & \underline{49.8} & 28.3 & 28.1 & 16.3 & 26.7 & \underline{15.5} & 26.3 & 18.5 & 18.1 & 26.7 & 40.4 & 25.2 & 38.6 & 28.6 & 28.5 \\
 & \textit{(Ours)} $3{\times}3$ median & \textbf{1.99} & \underline{2.71} & \underline{1.32} & \underline{3.36} & \textbf{1.98} & \textbf{1.91} & \textbf{30.0} & \textbf{50.9} & \textbf{22.9} & \textbf{50.1} & \textbf{33.2} & \textbf{32.4} & \textbf{19.9} & \textbf{28.2} & 14.7 & \textbf{28.7} & \textbf{20.4} & \textbf{20.4} & \textbf{32.7} & \textbf{51.0} & \textbf{29.4} & \textbf{50.4} & \textbf{34.0} & \textbf{33.8} \\
\midrule
\multirow{4}{*}{\rotatebox[origin=c]{90}{Aggregation}}
 & Mean & \underline{2.00} & 2.68 & \textbf{1.40} & 3.23 & 1.97 & \underline{1.91} & \underline{29.9} & 50.5 & 20.2 & 48.6 & \underline{33.3} & 32.2 & 18.2 & 25.0 & 13.4 & 25.2 & 19.3 & 18.8 & 30.3 & 43.5 & 25.9 & 42.3 & 32.0 & 31.3 \\
 & Median & 1.96 & 2.69 & 1.32 & \underline{3.29} & 1.93 & 1.87 & 29.8 & 50.2 & \textbf{24.7} & 49.0 & 32.6 & 32.0 & \textbf{20.0} & \textbf{28.4} & \textbf{16.0} & \underline{28.1} & \textbf{20.7} & \textbf{20.7} & \underline{31.9} & \underline{48.3} & \textbf{29.4} & \underline{48.1} & \underline{32.8} & \underline{32.8} \\
 & Rank-weighted mean & \textbf{2.01} & \textbf{2.75} & \underline{1.39} & 3.28 & \textbf{1.99} & \textbf{1.94} & \textbf{30.0} & \textbf{51.0} & 19.1 & \underline{49.1} & \textbf{33.5} & \underline{32.3} & 18.3 & 25.7 & 12.8 & 25.8 & 19.7 & 19.0 & 30.8 & 46.2 & \underline{26.0} & 44.1 & 32.7 & 32.0 \\
 & \textit{(Ours)} Rank-weighted median & 1.99 & \underline{2.71} & 1.32 & \textbf{3.36} & \underline{1.98} & \underline{1.91} & \textbf{30.0} & \underline{50.9} & \underline{22.9} & \textbf{50.1} & 33.2 & \textbf{32.4} & \underline{19.9} & \underline{28.2} & \underline{14.7} & \textbf{28.7} & \underline{20.4} & \underline{20.4} & \textbf{32.7} & \textbf{51.0} & \textbf{29.4} & \textbf{50.4} & \textbf{34.0} & \textbf{33.8} \\
\bottomrule
\end{tabular*}
\caption{Ablation on PASCAL VOC per model. Rows as in Tab.~\ref{tab:ablations} of the main paper. Each block varies one component and keeps the other two fixed. Best per block, model and metric in bold, second best underlined.}
\label{tab:app-ablation-models}
\end{table*}
\subsection{Understanding Model Outputs with \wish}
\label{app:model-outputs}

\begin{figure*}[t]
\centering
\includegraphics[width=\textwidth]{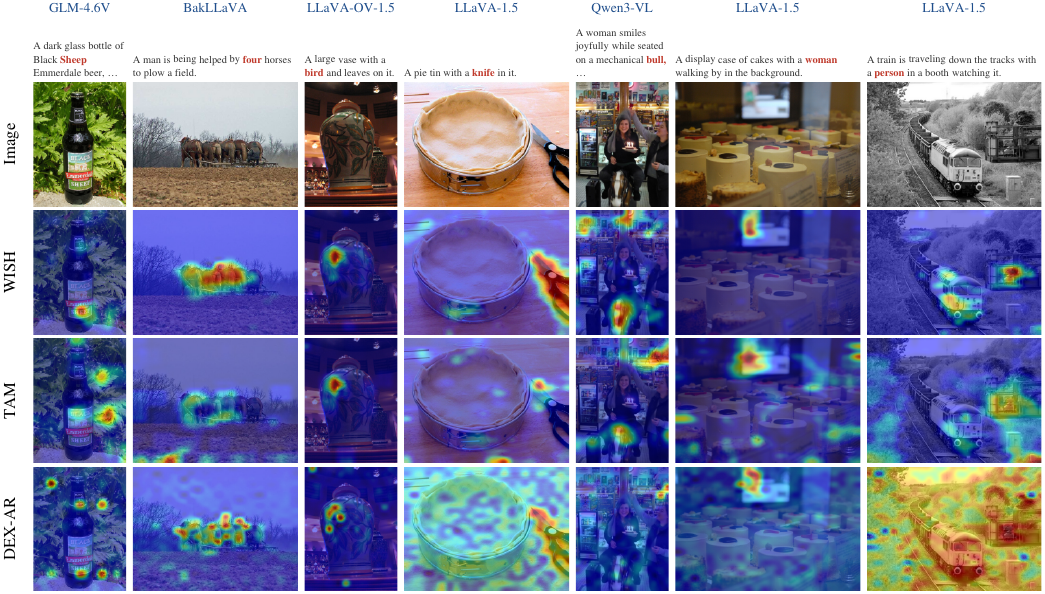}
\caption{\wish, TAM and DEX-AR maps of single words of model captions on PASCAL VOC (first two columns) and COCO
Caption (other columns). The model is named above each column, and the explained word is in red. From left to right:
reading text on a label, an undercount (four of six horses), a small painted detail, and four words for objects
that are not in the image.}
\label{fig:app-model-outputs}
\end{figure*}

Fig.~\ref{fig:app-model-outputs} uses \wish to inspect the visual evidence behind single words of model captions next to the maps of TAM and DEX-AR. 
The examples cover text in the image, a counting error, small details and confusions between objects.

\noindent\textbf{Reading text.} GLM-4.6V names the beer ``Black Sheep''. \wish's map of ``Sheep'' lies on the word
SHEEP printed on the label. TAM highlights the leaves next to the bottle, and DEX-AR gives scattered spots.

\noindent\textbf{Counting.} BakLLaVA counts ``four horses'' instead of six. \wish's map of ``four'' is strongest in
the middle of the team, where several horses overlap, and stays low on the horse that is hidden behind the person.
This fits a count that merges overlapping horses and misses the hidden one.

\noindent\textbf{Small details.} LLaVA-OV-1.5 describes ``a large vase with a bird and leaves on it''. \wish points at
the small bird painted on the vase, a detail that is easy to miss. TAM finds the same spot.

\noindent\textbf{Confusions between objects.} LLaVA-1.5 describes ``a pie tin with a knife in it''. The image contains
scissors and no knife, and \wish's map of ``knife'' lies on the scissors. DEX-AR also reaches the scissors, while
TAM's map is faint. Qwen3-VL calls the ride ``a mechanical bull''. \wish's map of ``bull'' lies on the toy horse
that the woman sits on, which TAM and DEX-AR cover only weakly. For a display case of cakes, LLaVA-1.5 adds ``a woman
walking by in the background''. \wish's map of ``woman'' lies at the top of the image. TAM and
DEX-AR highlight the same region. LLaVA-1.5 writes that ``a person in a booth'' watches the train. \wish and TAM point
at a structure next to the track, which the model took for a person in a booth.

\end{document}